\documentclass[sigconf]{acmart}

\usepackage{bm}
\usepackage{subfigure}
\usepackage{multirow}
\usepackage{xspace}
\usepackage{listings}
\usepackage{xcolor}
\usepackage{makecell}
\usepackage{CJKutf8}
\usepackage{algorithm}
\usepackage{algorithmic}
\newcommand{\sys}{\texttt{TransBO}\xspace}
\newcommand{\tabincell}[2]{\begin{tabular}{@{}#1@{}}#2\end{tabular}}
\newcommand{\para}[1]{{\vspace{2pt} \bf \noindent #1 \hspace{1pt}}}
\definecolor{codegray}{rgb}{0.5,0.5,0.5}

\AtBeginDocument{%
  \providecommand\BibTeX{{%
    \normalfont B\kern-0.5em{\scshape i\kern-0.25em b}\kern-0.8em\TeX}}}

\setcopyright{acmcopyright}
\copyrightyear{2022}
\acmYear{2022}
\setcopyright{acmcopyright}\acmConference[KDD '22]{Proceedings of the 28th ACM
SIGKDD Conference on Knowledge Discovery and Data Mining}{August 14--18,
2022}{Washington, DC, USA}
\acmBooktitle{Proceedings of the 28th ACM SIGKDD Conference on Knowledge
Discovery and Data Mining (KDD '22), August 14--18, 2022, Washington, DC, USA}
\acmPrice{15.00}
\acmDOI{10.1145/3534678.3539255}
\acmISBN{978-1-4503-9385-0/22/08}

\begin{document}

\title{\sys: Hyperparameter Optimization via Two-Phase Transfer Learning}

\author{Yang Li$^{\dagger\mathsection}$,
Yu Shen$^\dagger$, 
Huaijun Jiang$^\dagger$, 
Wentao Zhang$^\dagger$,
Zhi Yang$^\dagger$,
Ce Zhang$^\ddagger$, 
Bin Cui$^{\dagger\diamond}$
}

\affiliation{
$^\dagger$
School of CS \& Key Laboratory of High Confidence Software Technologies (MOE), Peking University\country{China}
}

\affiliation{
$^\mathsection$Data Platform, TEG, Tencent Inc.\country{China}
}

\affiliation{
$^\ddagger$Department of Computer Science, Systems Group, ETH Zurich\country{Switzerland}
}

\affiliation{
$^\diamond$Institute of Computational Social Science, Peking University (Qingdao)\country{China}
}

\affiliation{
$^\dagger$\{liyang.cs, shenyu, wentao.zhang,  jianghuaijun, yangzhi, bin.cui\}@pku.edu.cn ~~~~~~
$^\ddagger$ce.zhang@inf.ethz.ch\country{}
}\country{}

\renewcommand{\authors}{Yang Li, Yu Shen, Huaijun Jiang, Wentao Zhang, Zhi Yang, Ce Zhang, Bin Cui}
\renewcommand{\shortauthors}{Li et al.}

\begin{abstract}
With the extensive applications of machine learning models, automatic hyperparameter optimization (HPO) has become increasingly important.
Motivated by the tuning behaviors of human experts, it is intuitive to leverage auxiliary knowledge from past HPO tasks to accelerate the current HPO task.
In this paper, we propose \sys, a novel two-phase transfer learning framework for HPO, which can deal with the complementary nature among source tasks and dynamics during knowledge aggregation issues simultaneously.
This framework extracts and aggregates source and target knowledge jointly and adaptively, where the weights can be learned in a principled manner.
The extensive experiments, including static and dynamic transfer learning settings and neural architecture search, demonstrate the superiority of \sys over the state-of-the-arts.
\end{abstract}




\maketitle

\section{Introduction}
Machine learning (ML) models have been extensively applied in many fields such as recommendation, computer vision, financial market analysis, etc~\cite{hinton2012deep,he2016deep,goodfellow2016deep,he2017neural,devlin2018bert,henrique2019literature}.
However, the performance of ML models heavily depends on the choice of hyperparameter configurations (e.g., learning rate or the number of hidden layers in a deep neural network).
As a result, automatically tuning the hyperparameters has attracted lots of interest from both academia and industry~\cite{quanming2018taking}.
Bayesian optimization (BO) is one of the most prevailing frameworks for automatic hyperparameter optimization (HPO)~\cite{hutter2011sequential, bergstra2011algorithms, snoek2012practical}.
The main idea of BO is to use a surrogate model, typically a Gaussian Process (GP)~\cite{rasmussen2004gaussian}, to describe the relationship between a hyperparameter configuration and its performance (e.g., validation error), and then utilize this surrogate to determine the next configuration to evaluate by optimizing an acquisition function that balances exploration and exploitation. 

Hyperparameter optimization (HPO) is often a computationally-intensive process as one often needs to choose and evaluate hyperparameter configurations by training and validating the corresponding ML models.
However, for ML models that are computationally expensive to train (e.g., deep learning models or models trained on large-scale datasets), vanilla Bayesian optimization (BO) suffers from the low-efficiency issue~\cite{falkner2018bohb,li-mfeshb,li2021volcanoml} due to insufficient configuration evaluations within a limited budget.

{\bf (Opportunities) }
Production ML models usually need to be constantly re-tuned as new task / dataset comes or underlying code bases are updated, e.g., in the AutoML applications. 
The optimal hyperparameters may also change as the data and code change, and so should be frequently re-optimized. 
Although they may change significantly, the region of good or bad configurations may still share some correlation with those of previous tasks~\cite{yogatama2014efficient}, and this provides the opportunities towards a faster hyperparameter search.
Therefore, \textit{we can leverage the tuning results (i.e., observations) from previous HPO tasks (source tasks) to speed up the current HPO task (target task) via a transfer learning-based framework.}

{\bf (Challenges) }
The transfer learning for HPO consists of two key 
operations: \emph{extracting} source knowledge from previous HPO 
tasks, and \emph{aggregating} and \emph{transfering}
these knowledge to a target domain.
To fully unleash the potential of TL, we need to address two main challenges when performing the above operations:
1) {\em The Complementary Nature among Source Tasks. }
Different source tasks are often complementary and thus require us to treat them in a joint and cooperative manner.
Ignoring the synergy of multiple source tasks might lead to the loss of auxiliary knowledge.
2) {\em Dynamics during Knowledge Aggregation. }
At the beginning of HPO, the knowledge from the source tasks could bring benefits due to the scarcity of observations on the target task. 
However, as the tuning process proceeds, 
we should shift the focus to the target task.
Since the target task gets more observations, transferring from source tasks might not be necessary anymore considering the bias and noises in the source tasks (i.e., negative transfer~\cite{pan2010a}).
Existing methods~\cite{wistuba2016two,schilling2016scalable,feurer2018scalable} 
have been focusing on these two challenges. However, none of them considers both simultaneously.
This motivates our work, which aims at developing a transfer learning framework that could 1) extract source knowledge in a {\em cooperative} manner, and 2) transfer the auxiliary knowledge in an {\em adaptive} way.

In this paper, we propose \sys, a novel two-phase transfer learning framework for automatic HPO
that tries to address the above two challenges simultaneously.
\sys works under the umbrella of Bayesian optimization and designs a transfer learning (TL) surrogate to guide the HPO process.
This framework decouples the process of knowledge transfer into two phases and considers the knowledge extraction and knowledge aggregation separately in each phase (See Figure~\ref{framework}).
In Phase one, \sys builds a source surrogate that extracts and combines useful knowledge across multiple source tasks.  
In Phase two, \sys integrates the source surrogate (in Phase one) and the target surrogate to construct the final surrogate, which we refer to as the transfer learning surrogate.
To maximize the generalization of the transfer learning surrogate, we adopt the cross-validation mechanism to learn the transfer learning surrogate in a principled manner.
Moreover, instead of combining base surrogates with independent weights, \sys can learn the optimal aggregation weights for base surrogates jointly.
To this end, we propose to learn the weights in each phase by solving a constrained optimization problem with a differentiable ranking loss function.

The empirical results of static TL scenarios showcase the stability and effectiveness of \sys compared with state-of-the-art TL methods for HPO. In dynamic TL scenarios that are close to real-world applications, \sys obtains strong performance -- the top-2 results on 22.25 out of 30 tuning tasks (Practicality). 
In addition, when applying \sys to neural architecture search (NAS), it achieves more than 5$\times$ speedups than the state-of-the-art NAS approaches (Universality).

{\bf (Contributions )}
In this work, our main contributions are summarized as follows:
\begin{itemize}
    \item We present a novel two-phase transfer learning framework for HPO --- \sys, which could address the aforementioned challenges simultaneously.
    \item We formulate the learning of this two-phase framework into constrained optimization problems. By solving these problems, \sys could extract and aggregate the source and target knowledge in a joint and adaptive manner.
    \item To facilitate transfer learning research for HPO, we create and publish a large-scale benchmark, which takes more than 200K CPU hours and involves more than 1.8 million model evaluations.
    \item The extensive experiments, including static and dynamic TL settings and neural architecture search, demonstrate the superiority of \sys over state-of-the-art methods.
\end{itemize}

\section{Related Work}
Bayesian optimization (BO) has been successfully applied to hyperparameter optimization (HPO)~\cite{bischl2021hyperparameter,li2020efficient,openbox,li2022hyper}.
For ML models that are computationally expensive to train (e.g., deep learning models or models trained on large datasets), BO methods~\cite{hutter2011sequential,bergstra2011algorithms,snoek2012practical} suffer from the low-efficiency issue due to insufficient configuration evaluations within a limited budget.
To speed up HPO of ML algorithms with limited trials, recent BO methods extend the traditional black-box assumption by exploiting cheaper fidelities from the current task \cite{klein2017fast, swersky2014freeze,kandasamy2017multi, klein2016learning,poloczek2017multi, falkner2018bohb,li-mfeshb,li2022hyper}. Orthogonal to these methods, we focus on borrowing strength from previously finished tasks to accelerate the HPO of the current task.

Transfer learning (TL) methods for HPO aim to leverage auxiliary knowledge from previous tasks to achieve faster optimization on the target task. 
One common way is to learn surrogate models from past tuning history and use them to guide the search of hyperparameters.
For instance, several methods learn all available information from both source and target tasks in a single surrogate, and make the data comparable through a transfer stacking ensemble~\cite{pardoe2010boosting}, a ranking algorithm~\cite{bardenet2013collaborative}, multi-task GPs~\cite{swersky2013multi}, a mixed kernel GP~\cite{yogatama2014efficient}, the GP noisy model~\cite{joy2016flexible}, a multi-layer perceptron with Bayesian linear regression heads~\cite{snoek2015scalable,perrone2018scalable} or replace GP with Bayesian neural networks~\cite{springenberg2016bayesian}.
SGPR~\cite{golovin2017google} and SMFO~\cite{wistuba2015sequential} utilize the knowledge from all source tasks equally and thus suffer from performance deterioration when the knowledge of source tasks is not applicable to the target task.
FMLP~\cite{schilling2015hyperparameter} uses multi-layer perceptrons as the surrogate model that learns the interaction between hyperparameters and datasets.
SCoT~\cite{bardenet2013collaborative} and MKL-GP~\cite{yogatama2014efficient} fit a GP-based surrogate on merged observations from both source tasks and target task.
To distinguish the varied performance of the same configuration on different tasks, the two methods use the meta-features of datasets to represent the tasks;
while the meta-features are often unavailable for broad classes of HPO problems~\cite{feurer2018scalable}.
Due to the high computational complexity of GP ($\mathcal{O}(n^3)$), it is difficult for these methods to scale to a large number of source tasks and trials (scalability bottleneck). 

To improve scalability, recent methods adopt the ensemble framework to conduct TL for HPO, where they train a base surrogate on each source task and the target task respectively and then combine all base surrogates into an ensemble surrogate with different weights. 
This framework ignores the two aforementioned issues and uses the {\em independent} weights.
POGPE \cite{schilling2016scalable} sets the weights of base surrogates to constants.
TST \cite{wistuba2016two} linearly combines the base surrogates with a Nadaraya-Watson kernel weighting by defining a distance metric across tasks; the weights are calculated by using either meta-features (TST-M) or pairwise hyperparameter configuration rankings (TST-R).
RGPE \cite{feurer2018scalable} uses the probability that the base surrogate has the lowest ranking loss on the target task to estimate the weights.
Instead of resorting to heuristics, \sys propose to learn the joint weights in a principled way.

Warm-starting methods~\cite{lindauer2018warmstarting,kim2017learning} select several initial hyperparameter configurations as the start points of search procedures. 
\citet{salinas2020quantile} deal with the heterogeneous scale between tasks with the Gaussian Copula Process.
ABRAC~\cite{horvath2021hyperparameter} proposes a multi-task BO method with adaptive complexity to prevent over-fitting on scarce target observations.
TNP~\cite{wei2021meta} applies the neural process to jointly transfer surrogates, parameters, and initial configurations.
Recently, transferring search space has become another way for applying transfer learning in HPO. ~\citet{wistuba2015hyperparameter} prune the bad regions of search space according to the results from previous tasks. 
This method suffers from the complexity of obtaining meta-features and relies on some other parameters to construct a GP model. 
On that basis, ~\citet{NIPS2019_9438} propose to utilize previous tasks to design a sub-region of the entire search space for the new task. 
While sharing some common spirits, these methods are orthogonal and complementary to our surrogate transfer method introduced in this paper.

In addition, our proposed two-phase framework inherits the advantages of the bi-level optimization~\cite{bennett2008bilevel}.
While previous methods in the literature focus on different tasks (e.g., evolutionary computation~\cite{sinha2017review}), to the best of our knowledge, \sys is the first method that adopts the concept of bi-level optimization into hyperparameter transfer learning.

\section{Bayesian Hyperparameter Optimization}
The HPO of ML algorithms can be modeled as a black-box optimization problem. 
The goal is to find $argmin_{\bm{x} \in \mathcal{X}}f(\bm{x})$ in hyperparameter space $\mathcal{X}$, where $f(\bm{x})$ is the ML model's performance metric (e.g., validation error) corresponding to the configuration $\bm{x}$. 
Due to the intrinsic randomness of most ML algorithms, we evaluate configuration $\bm{x}$ and can only get its noisy result $y = f(\bm{x}) + \epsilon$ with $\epsilon \sim \mathcal{N}(0, \sigma^2)$. 

{\bf Bayesian optimization (BO)} is a model-based framework for HPO. 
BO first fits a probabilistic surrogate model $M:p(f|D)$ on the already observed instances $D=\{(\bm{x}_1, y_1),...,(\bm{x}_{n-1}, y_{n-1})\}$.
In the $n$-th iteration, BO iterates the following steps: 1) use surrogate $M$ to select a promising configuration $\bm{x}_n$ that maximizes the acquisition function $\bm{x}_{n}=\arg\max_{\bm{x} \in \mathcal{X}}a(\bm{x}; M)$, where the acquisition function is to balance the exploration and exploitation trade-off; 2) evaluate this point to get its performance $y_n$, and add the new observation $(\bm{x}_{n}, y_{n})$ to $D$; 3) refit $M$ on the augmented $D$. 
Expected Improvement (EI)~\cite{jones1998efficient} is a common acquisition function defined as follows:
\begin{equation}
\label{eq_ei}
a(\bm{x}; M)=\int_{-\infty}^{\infty} \max(y^{\ast}-y, 0)p_{M}(y|\bm{x})dy,
\end{equation}
where $M$ is the surrogate and $y^{\ast}=\min\{y_1, ..., y_n\}$. 
By maximizing this EI function $a(\bm{x}; M)$ over $\mathcal{X}$, BO methods can find a configuration to evaluate for each iteration.

\begin{figure}[!t]
\begin{center}
\centerline{\includegraphics[width=\columnwidth]{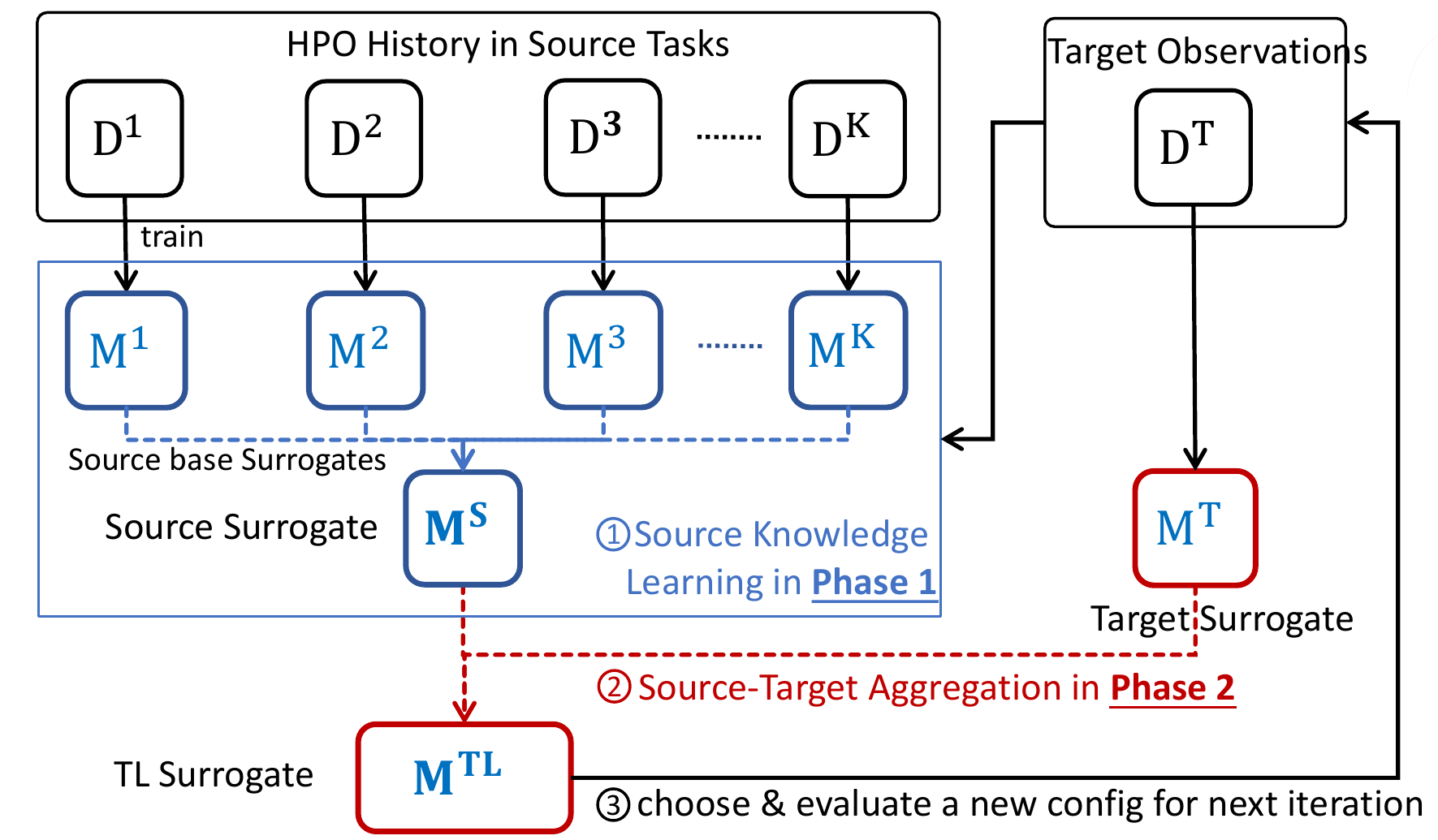}}
\caption{Two-Phase Transfer Learning Framework.}
\vspace{-2em}
\label{framework}
\end{center}
\end{figure}

\section{The Proposed Method}
\label{sec4}
In this section, we present \sys, a two-phase transfer learning (TL) framework for HPO. Before diving into the proposed framework, we first introduce the notations and settings for TL. Then we describe \sys in details and end the section with discussions about its advantages.

\para{Basic Notations and Settings. }
As illustrated in Figure~\ref{framework},
we denote observations from $K+1$ tasks as $D^1$, ..., $D^K$
for $K$ source tasks and $D^T$ for the target task. 
The $i$-th source task has $n_i$ configuration observations: $D^i=\{(\bm{x}_j^i, y_j^i)\}_{j=1}^{n_i}$ with $i=1,2,...,K$, which are obtained from previous tuning procedures.
For the target task, after completing $t$ iterations (trials), the observations in the target task are: $D^T=\{(\bm{x}_j^T, y_j^T)\}_{j=1}^{t}$.

Before optimization, we train a base surrogate model for the $i$-th source task, denoted by $M^i$.
Each base surrogate $M^i$ can be fitted on $D^i$ in advance (offline), and the target surrogate $M^T$ is trained on $D^T$ on the fly.
Since the configuration performance $y$s in each $D^i$ and $D^T$ may have different numerical ranges, we standardize the $y$s in each task by removing the mean and scaling to unit variance. 
For a hyperparameter configuration $\bm{x}_j$, each base surrogate $M^i$ outputs a posterior predictive distribution at $\bm{x}_j$, that's, $M^i(\bm{x}_j) \sim \mathcal{N}(\mu_{M^i}(\bm{x}_j), \sigma^2_{M^i}(\bm{x}_j))$. For brevity, we denote the mean of this prediction at $\bm{x}_j$ as $M^i(\bm{x}_j)=\mu_{M^i}(\bm{x}_j)$.

\subsection{Overview}
\sys aims to build a transfer learning surrogate model $M^{TL}$ on the target task, which outputs a more accurate prediction for each configuration by borrowing strength from the source tasks.
The cornerstone of \sys is to decouple the combination of $K+1$ base surrogates with a novel two-phase framework:

{\em Phase 1.}
To leverage the complementary nature among source tasks, \sys first linearly combines all source base surrogates into a single source surrogate with the weights ${\bf w}$:
\begin{equation}
    M^S = \texttt{agg}(\{M^1,...,M^K\}; {\bf w}).
\nonumber
\end{equation}
In this phase, the useful source knowledge from each source task is extracted and integrated into the source surrogate in a joint and cooperative manner.

{\em Phase 2.}
To support dynamics-aware knowledge aggregation, \sys further combines the aggregated source surrogate with the target surrogate $M^T$ via weights ${\bf p}$ in an adaptive manner, where $M^T$ is trained on the target observations $D^{T}$:
\begin{equation}
M^{TL} = \texttt{agg}(\{M^S, M^T\}; {\bf p}).
\nonumber
\end{equation}

Such joint and adaptive knowledge transfer in two phases guarantees the efficiency and effectiveness of the final TL surrogate $M^{TL}$ in extracting and integrating the source and target knowledge. 
To maximize the generalization ability of $M^{TL}$, the two-phase framework further learns the parameters ${\bf w}$ and ${\bf p}$ in a principled and automatic manner by solving the constrained optimization problems.
In the following, we describe the parameter learning and aggregation method.

\subsection{Parameter Learning in Two-Phase Framework}
\label{sec_tp_learning}
Notice that
${\bf w}$ and ${\bf p}$ play different 
roles --- ${\bf w}$ combines $K$ source base surrogates to best fit the target observations, while ${\bf p}$ balances between two surrogates $M^S$ and $M^T$. 
The objective of \sys is to maximize the generalization performance of $M^{TL}$. 
To obtain ${\bf w}$, we use the target observations $D^T$ to maximize the performance of source surrogate $M^{S}$.
However, if we learn the parameter ${\bf p}$ of $M^{TL}$ on $D^T$ by using the $M^{S}$ and $M^T$, where $M^{S}$ and $M^{T}$ are trained on $D^T$ directly, the learning process becomes an estimation of in-sample error and can not reflect the generalization of the final surrogate $M^{TL}$.
To address this issue, we adopt the cross-validation mechanism to maximize the generalization ability of $M^{TL}$ when learning ${\bf p}$.
In the following, we first describe the general procedure to learn a surrogate $M^{S}$ on given observations $D$ (instead of $D^T$), and then introduce the method to learn the parameters $\bf w$ and $\bf p$, respectively.

\para{General Procedure: Fitting $M^S$ on Given Observations $D$.}
Our strategy is to obtain the source surrogate $M^S$ as a weighted combination of the predictions of source base surrogates $\{M^1,...,M^K\}$:
\begin{equation}
\label{fS}
M^S(\bm{x}) = \sum_{i=1}^K{w}_{i}M^i(\bm{x}),
\end{equation}
where $\sum_{i}{w}_i = 1$ and ${w}_i \in [0, 1]$.
Intuitively, the weight $w_i$ reflects the quality of knowledge extracted from the corresponding source tasks. 
Instead of calculating weights independently, which may ignore the complementary nature among source tasks, we propose to combine source base surrogates $M^i$s in a joint and supervised manner, which reveals their cooperative contributions to $M^S$. 

To derive $M^S$ in a principled way, we use a differentiable pairwise ranking loss function to measure the fitting error between the prediction of $M^S$ and the available observations $D$. 
In HPO, ranking loss is more appropriate than mean square error --- the actual values of predictions are not the most important, and we care more about the partial orders over the hyperparameter space, e.g., the location of the optimal configuration.
This ranking loss function is defined as follows:
\begin{equation}
\begin{aligned}
    & \mathbb{L}({\bf w}, M^{S}; D) = \frac{1}{n^2}\sum_{j=1}^{n}\sum_{k=1,y_j<y_k}^{n}\phi(M^S(\bm{x}_k) - M^S(\bm{x}_j)), \\
    & \phi(z) = log(1 + e^{-z}), \\
\end{aligned}
\label{ranking_loss}
\end{equation}
where $n$ is the number of observations in $D$, $y$ is the observed performance of configuration $\bm{x}$ in $D$, 
and the prediction of $M^S(\bm{x}_j)$ at configuration $\bm{x}_j$ is obtained by linearly combining the predictive mean of $M^i$ with a weight ${w}_i$, that's, $M^S(\bm{x}_j)=\sum_i { w}_iM^i(\bm{x}_j)$.

We further turn the learning of source surrogate $M^{S}$, i.e., the learning of ${\bf w}$, into the following constrained optimization problem:
\begin{equation}
    \label{eq:opt_source}
    \begin{aligned}
        & \underset{{\bf w}}{\text{minimize}}
        & & \mathbb{L}({\bf w}, M^{S}; D) \\
        & \text{s.t.}
        & & \bm{1}^\top{\bf w}=1, {\bf w}\ge\bm{0}, \\
    \end{aligned}
\end{equation}
where the objective is the ranking loss of $M^S$ on $D$.
This optimization objective is continuously differentiable, and concretely, it is twice continuously differentiable. 
So we can have the first derivative of the objective $\mathbb{L}$ as follows:
\begin{equation}
\begin{aligned}
    & \frac{\partial \mathbb{L}}{\partial {\bf w}} = \sum_{(j, k) \in \mathbb{P}}\frac{neg\_ez}{1 + neg\_ez} \ast (A_{[j]} - A_{[k]}), \\
    & neg\_ez = e^{(A_{[j]}{\bf w} - A_{[k]}{\bf w})}, \\
\end{aligned}
\label{der_loss}
\end{equation}
where $\mathbb{P}$ consists of pairs $(j, k)$ satisfying $y_j < y_k$, $A$ is the matrix formed by putting the predictions of $M^{1:K}$s together where the element at the $i$-th row and $j$-th column is $M^i(\bm{x}_j)$, and $A_{[j]}$ is the row vector in the $j$-th row of matrix $A$.
Furthermore, this optimization problem can be solved efficiently by applying many existing sequential quadratic programming (SQP) solvers ~\cite{10.1145/192115.192124}. 

\para{Learning Parameter ${\bf w}$.}
As stated previously, to maximize the (generalization) performance of $M^{S}$, we propose to learn the parameter ${\bf w}$ by fitting $M^{S}$ on the whole observations $D^T$.
In this way, the useful source knowledge from multiple source tasks can be fully extracted and integrated in a joint manner.
Therefore, the parameters {\bf w} can be obtained by calling the general procedure, i.e., solving the problem~\ref{eq:opt_source}, where the available observations $D$ are set to $D^T$.

\para{Learning Parameter ${\bf p}$.}
To reflect the generalization in $M^{TL}$, the parameter ${\bf p}$ is learned with the cross-validation mechanism. 
We first split $D^T$ into
$N_{cv}$ partitions: $D^T_1$, ..., $D^T_{N_{cv}}$ with $N_{cv}=5$. 
For each partition $i\in [1:N_{cv}]$, we first fit a partial surrogate $M^S_{-i}$ on the observations $D^T_{-i}$ with observations in the $i$-th partition removed from $D^T$, and the surrogate $M^S_{-i}$ is learned on $D^T_{-i}$ using the general procedure; in addition, we also fit a partial surrogate model $M^T_{-i}$ on $D^T_{-i}$ directly.
Then we combine the surrogates $M^S_{-i}$ and $M^T_{-i}$ linearly to obtain a $M_{-i}^{TL}$:
\begin{equation}
\begin{aligned}
    M^{TL}_{-i} = {p}^S M^S_{-i} + {p}^T M^T_{-i},\\
\end{aligned}
\end{equation}
where ${\bf p} =[{p}^S, {p}^T]$. 
Therefore, we can obtain $N_{cv}$ partial surrogates $M_{-i}^S$ and $M_{-i}^T$ with $i\in[1:N_{cv}]$.
Based on the differentiable pairwise ranking loss function in Eq.~\ref{ranking_loss}, the loss of $M_{-i}^{TL}$ on $D^{T}$ is defined as:
\begin{equation}
\begin{aligned}
    & \mathbb{L}_{cv}({\bf p}, M^{TL}_{-i}; D^T) = \frac{1}{n^2}\sum_{j=1}^{n}\sum_{k=1,y_j^T<y_k^T, k \in D^T_{i}}^{n}\phi(z), \\
    & \phi(z) = log(1 + e^{-z}), z=M^{TL}_{-i}(\bm{x}_k) - M^{TL}_{-H(j)}(\bm{x}_j)\\
\end{aligned}
\label{ranking_loss_cv}
\end{equation}
where $n$ is the number of observations in $D^T$, $y^T$ is the observed performance of configuration $\bm{x}^T$ in $D^T$, $H(j)$ indicates the partition id that configuration $\bm{x}_j$ belongs to, and the prediction of $M^{TL}_{-i}$ at configuration $\bm{x_k}$ is obtained by linearly combining the predictive mean of $M^S_{-i}$ and $M^{T}_{-i}$ with weight {\bf p}, that's, $M_{-i}^{TL}(\bm{x}_k)={p}^SM^{S}_{-i}(\bm{x}_k)+{p}^TM^T_{-i}(\bm{x}_k)$.
So the parameter {\bf p} can be learned by solving a similar constrained optimization problem on $D^T$:
\begin{equation}
    \label{eq:opt_target}
    \begin{aligned}
        & \underset{{\bf p}}{\text{minimize}}
        & & \sum_{i=1}^{N_{cv}} \mathbb{L}_{cv}({\bf p}, M^{TL}_{-i}; D^T) \\
        & \text{s.t.}
        & & \bm{1}^\top{\bf p}=1, {\bf p}\ge\bm{0}.\\
    \end{aligned}
\end{equation}
Following the solution introduced in problem~\ref{eq:opt_source}, the above optimization problem can be solved efficiently.

\para{Final TL Surrogate.} 
After ${\bf w}$ and ${\bf p}$ are obtained, as illustrated in Figure~\ref{framework}, we first combine the source base surrogates into the source surrogate $M^S$ with ${\bf w}$ (the Phase 1), and then integrate $M^{S}$ and $M^T$ with ${\bf p}$ to obtain the final TL surrogate $M^{TL}$ (the Phase 2). To ensure the surrogate $M^{TL}$ still works in the BO framework, it is required to be a GP.
How to obtain the unified posterior predictive mean and variance from multiple GPs (base surrogates) is still an open problem. 
As suggested by \cite{feurer2018scalable}, the linear combination of multiple base surrogates works well in practice. 
Therefore, we aggregate the base surrogates with linear combination.
That's, suppose there are $N_B$ GP-based surrogates, and each base surrogate $M^b$ has a weight $w_b$ with $b = 1, ..., N_B$, the combined prediction under the linear combination technique is give by: $\mu_{C}(\bm{x})=\sum_bw_b\mu_{b}(\bm{x})$ and $\sigma^2_{C}(\bm{x})=\sum_bw_b^2\sigma_b^2(\bm{x})$.

\para{Algorithm Summary}
At initialization, we set the weight of each source surrogate in ${\bf w}$ to $1/K$, and ${\bf p}=[1, 0]$ when the number of trials is insufficient for cross-validation.
Algorithm~\ref{algo:tptl_framework} illustrates the pseudo code of \sys.
In the $i$-th iteration, we first learn the weights ${\bf p}_i$ and ${\bf w}_i$ by solving two optimization problems (Lines 2-3).
Since we have the prior: as the HPO process of the target task proceeds, the target surrogate owns more and more knowledge about the objective function of the target task, therefore the weight of $M^{T}$ should increase gradually. 
To this end, we employ a \emph{max} operator, which enforces that the update of ${p}^{T}$ should be non-decreasing (Line 4).
Next, by using linear combination, we build the source surrogate $M^{S}$ with weight ${\bf w}_i$, and then construct the final TL surrogate $M^{TL}$ with ${\bf p}_i$ (Line 5).
Finally, \sys utilizes $M^{TL}$ to choose a promising configuration to evaluate, and refit the target surrogate on the augmented observation (the BO framework, Lines 6-7).

\begin{algorithm}[tb]
  \small
  \caption{The \sys Framework.}
  \label{algo:tptl_framework}
  \textbf{Input}: maximum number of trials $N^{T}$, observations from $K$ source tasks: $D^{1:K}$, and config. space $\mathcal{X}$.
  \begin{algorithmic}[1]
    \FOR{$i \in \{1, 2, ..., N^{T}\}$}
      \STATE Calculate the weight ${\bf w}_i$ in $M^{S}$ by solving~(\ref{eq:opt_source}).
      \STATE Calculate the weight ${\bf p}_i$ in $M^{TL}$ by solving~(\ref{eq:opt_target}).
      \STATE Employ non-decreasing prior on ${p}^{T}$: $p^{T}_i = \operatorname{max}(p^{T}_i, p^{T}_{i-1})$.
      \STATE Build $M^{S}$, $M^{TL}$ with weights ${\bf w}_i$ and ${\bf p}_i$, respectively.
      \STATE Sample a large number of configurations randomly from $\mathcal{X}$, compute their acquisition values according to the EI criterion in Eq.\ref{eq_ei}, where $M = M^{TL}$, and choose the configuration $\bm{x}_i = \operatorname{argmax}_{x\in\mathcal{X}}ei(x, M^{TL})$.
      \STATE Evaluate $\bm{x}_i$ and get its performance $y_i$, augment observations $D^{T}$ with $(\bm{x_i}, y_i)$ and refit $M^{T}$ on the augmented $D^{T}$.
    \ENDFOR
  \STATE \textbf{return} the best configuration in $D^T$.
\end{algorithmic}
\end{algorithm}

\subsection{Discussion: Advantages of \sys}
To our knowledge, \sys is the first method that conducts transfer learning for HPO in a supervised manner, instead of resorting to some heuristics.
In addition, this method owns the following desirable properties simultaneously.
1) \textbf{Practicality.} 
A practical HPO method should be insensitive to its hyperparameters, and do not depend on meta-features. 
The goal of HPO is to optimize the ML hyperparameters automatically while having extra (or sensitive) hyperparameters itself actually violates its principle. 
In addition, many datasets, including image and text data, lack appropriate meta-features to represent the dataset~\cite{wistuba2016two,schilling2015hyperparameter,feurer2018scalable}.
The construction of TL surrogate in \sys is insensitive to its hyperparameters and does not require meta-features.
2)~\textbf{Universality.} The 1st property enable \sys to be a general transfer learning framework for Black-box optimizations, e.g., experimental design~\cite{NEURIPS2019_d55cbf21}, neural architecture search~\cite{dudziak2020brp}, etc; we include an experiment to evaluate \sys on the NAS task in the section of experiment).
3)~\textbf{Scalability.} Compared with the methods that combine $k$ source tasks with $n$ trials into a single surrogate ($O(k^3n^3)$), \sys has a much lower complexity $O(kn^3)$, which means that \sys could scale to a large number of tasks and trials easily. 
4)~\textbf{Theoretical Discussion.} 
\sys also provides theoretical discussions about preventing the performance deterioration (negative transfer). 
Base on cross-validation and the non-decreasing constraint, {\em the performance of \sys, given sufficient trials, will be no worse than the method without transfer learning}, while the other methods cannot have this (See Appendix~\ref{converge_analysis} for more details).

\section{Experiments and Results}
\label{sec:exp_sec}
In this section, we evaluate \sys from three perspectives: 1) stability and effectiveness on static TL tasks, 2) practicality on real-world dynamic TL tasks, and 3) universality when conducting neural architecture search.

\begin{figure*}[htb]
	\centering
	\subfigure[Random Forest]{
		\scalebox{0.23}{
			\includegraphics[width=1\linewidth]{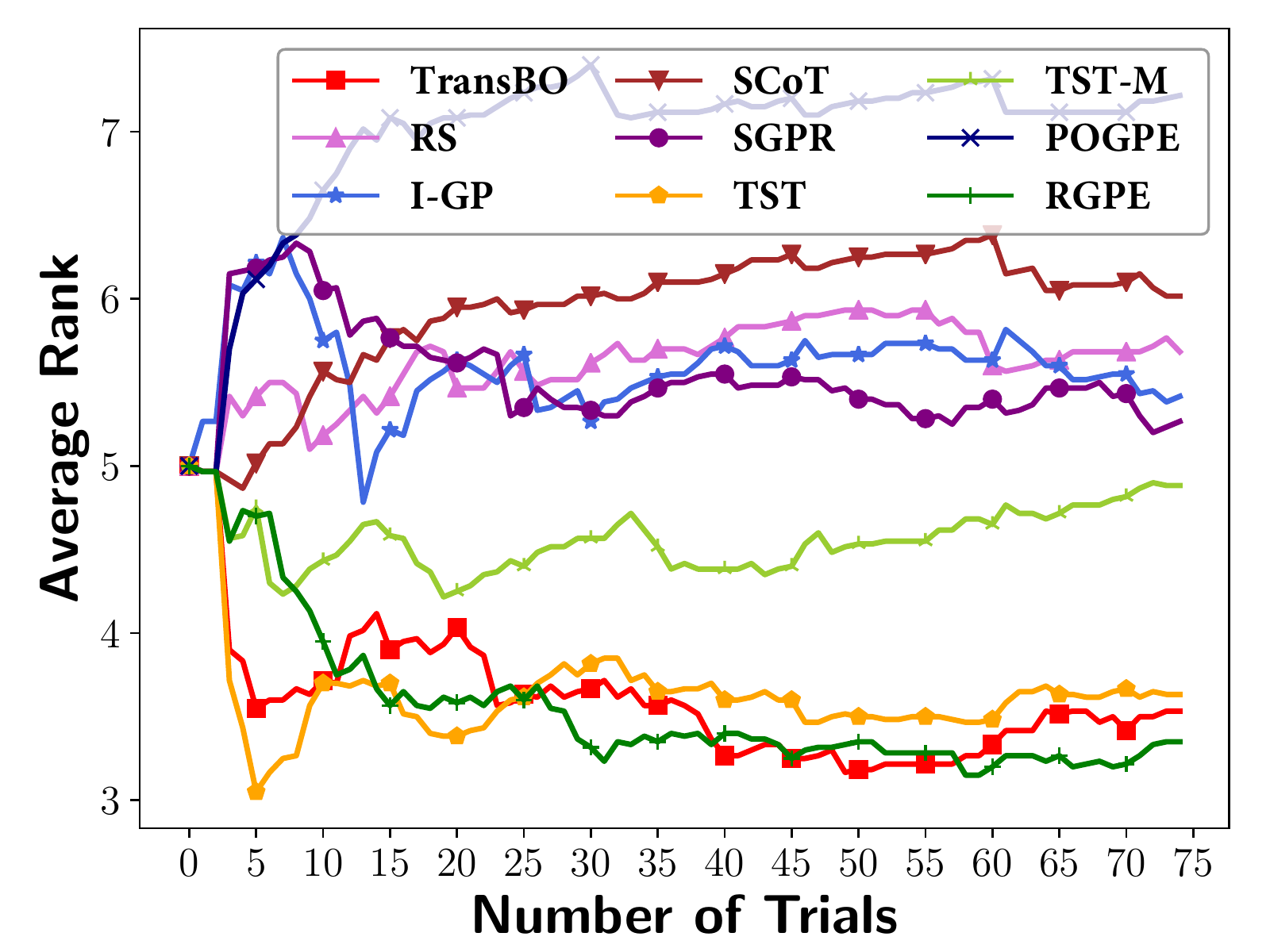}
			\label{offline_exp1_rf}
	}}
	\subfigure[LightGBM]{
		\scalebox{0.23}{
			\includegraphics[width=1\linewidth]{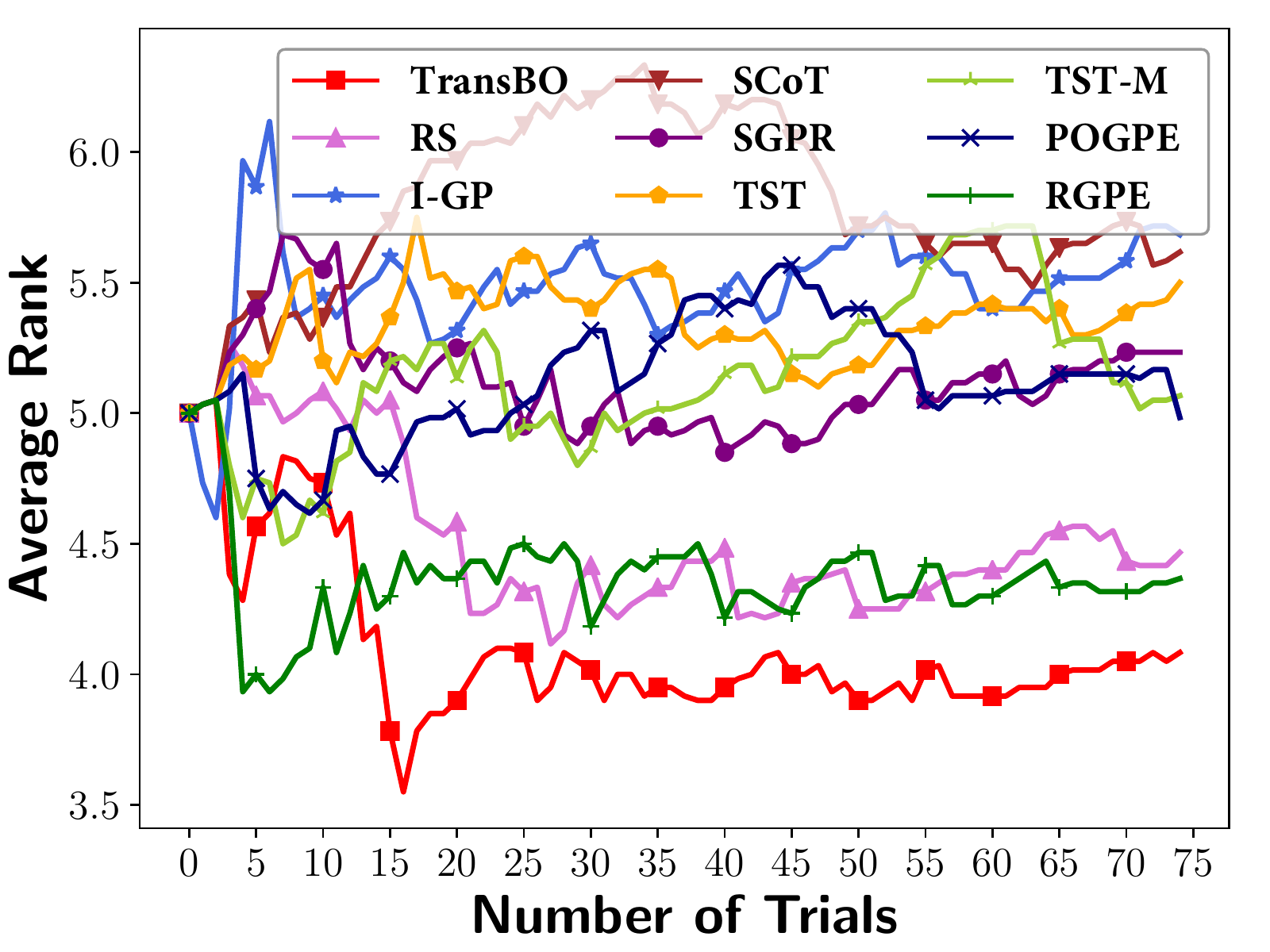}
			\label{offline_exp1_lgb}
	}}
	\subfigure[Adaboost]{
		\scalebox{0.23}{
			\includegraphics[width=1\linewidth]{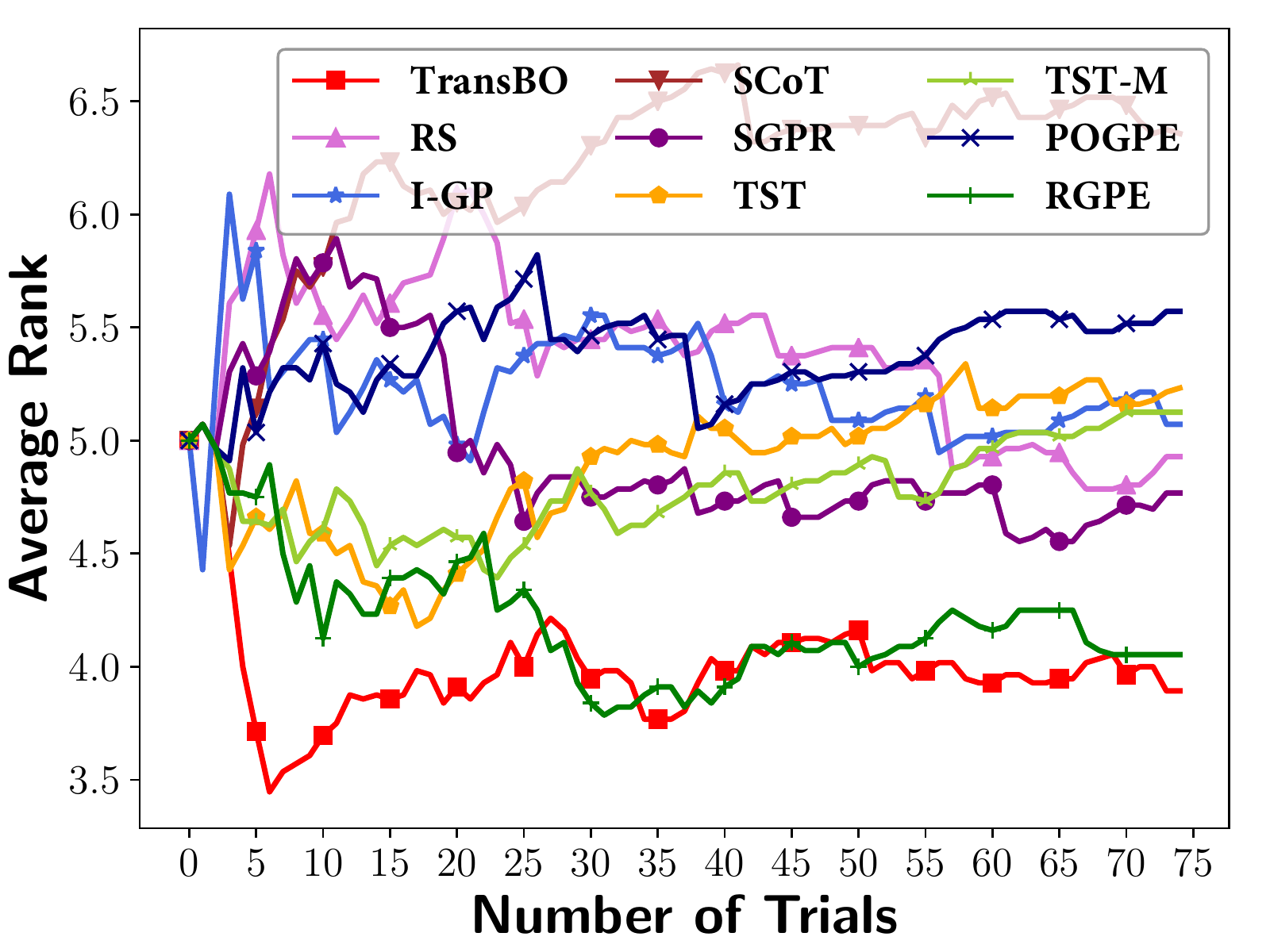}
			\label{offline_exp1_adb}
	}}
    \subfigure[Extra Trees]{
		\scalebox{0.23}{
			\includegraphics[width=1\linewidth]{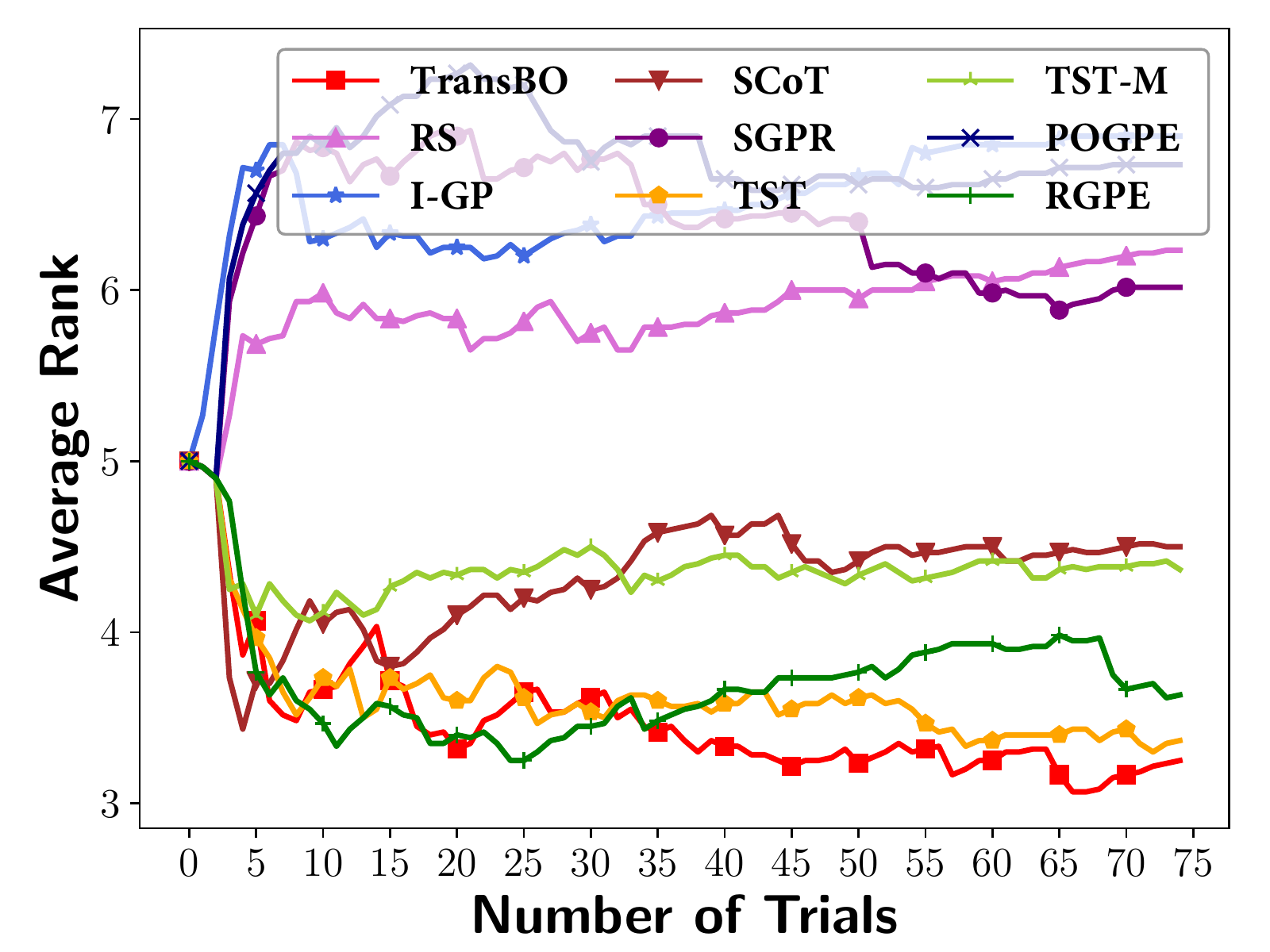}
			\label{offline_exp1_ext}
	}}
	\caption{Static TL results for four algorithms with $N_{task} = 29$ source tasks.}
  \label{offline_exp1}
\end{figure*}

\begin{figure*}[htb]
	\centering
	\subfigure[Random Forest]{
		\scalebox{0.23}{
		\includegraphics[width=1\linewidth]{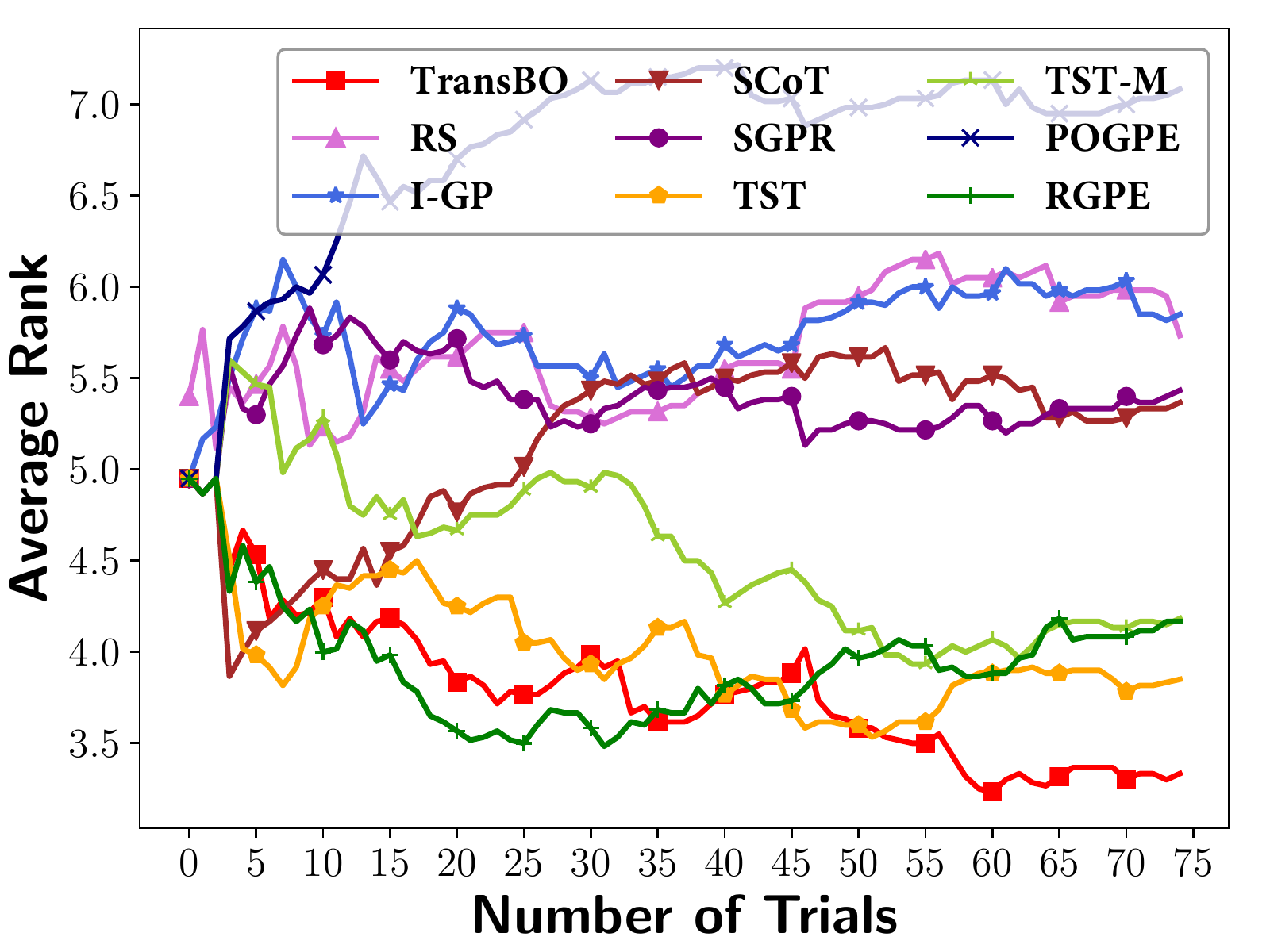}
			\label{offline_exp2_rf}
	}}
	\subfigure[LightGBM]{
		\scalebox{0.23}{
			\includegraphics[width=1\linewidth]{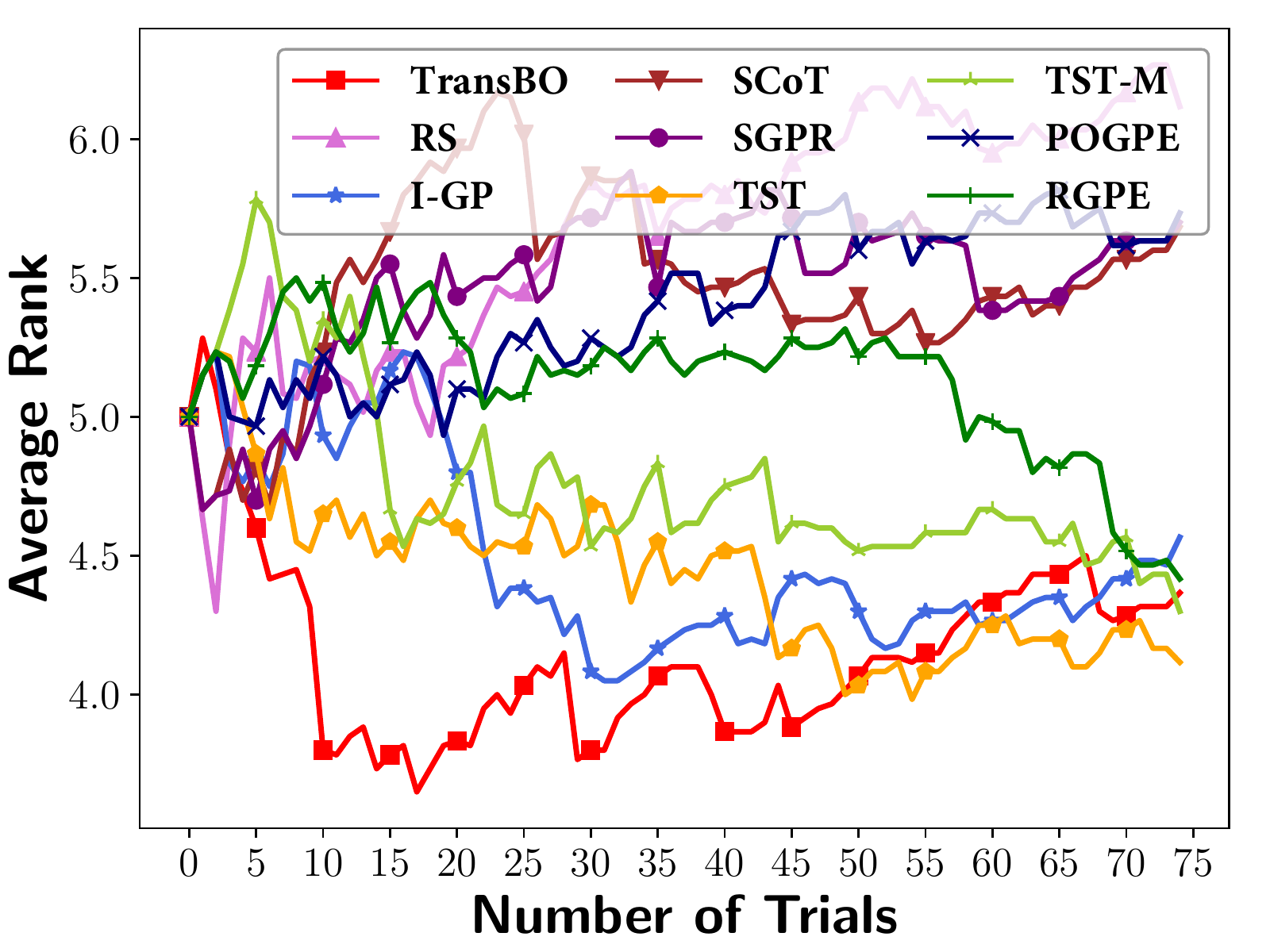}
			\label{offline_exp2_lgb}
	}}
	\subfigure[Adaboost]{
		\scalebox{0.23}{
			\includegraphics[width=1\linewidth]{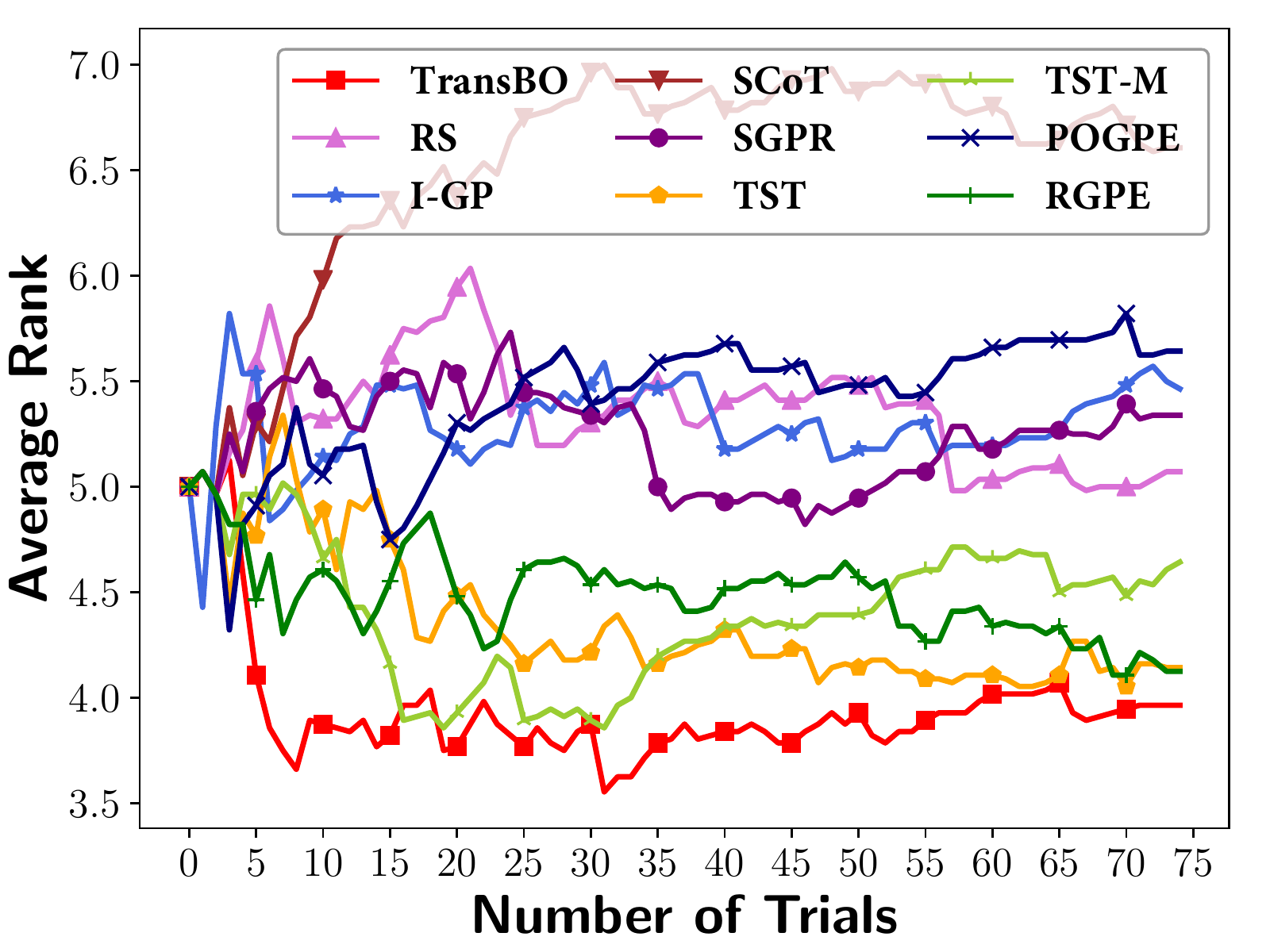}
			\label{offline_exp2_adb}
	}}
    \subfigure[Extra Trees]{
		\scalebox{0.23}{
			\includegraphics[width=1\linewidth]{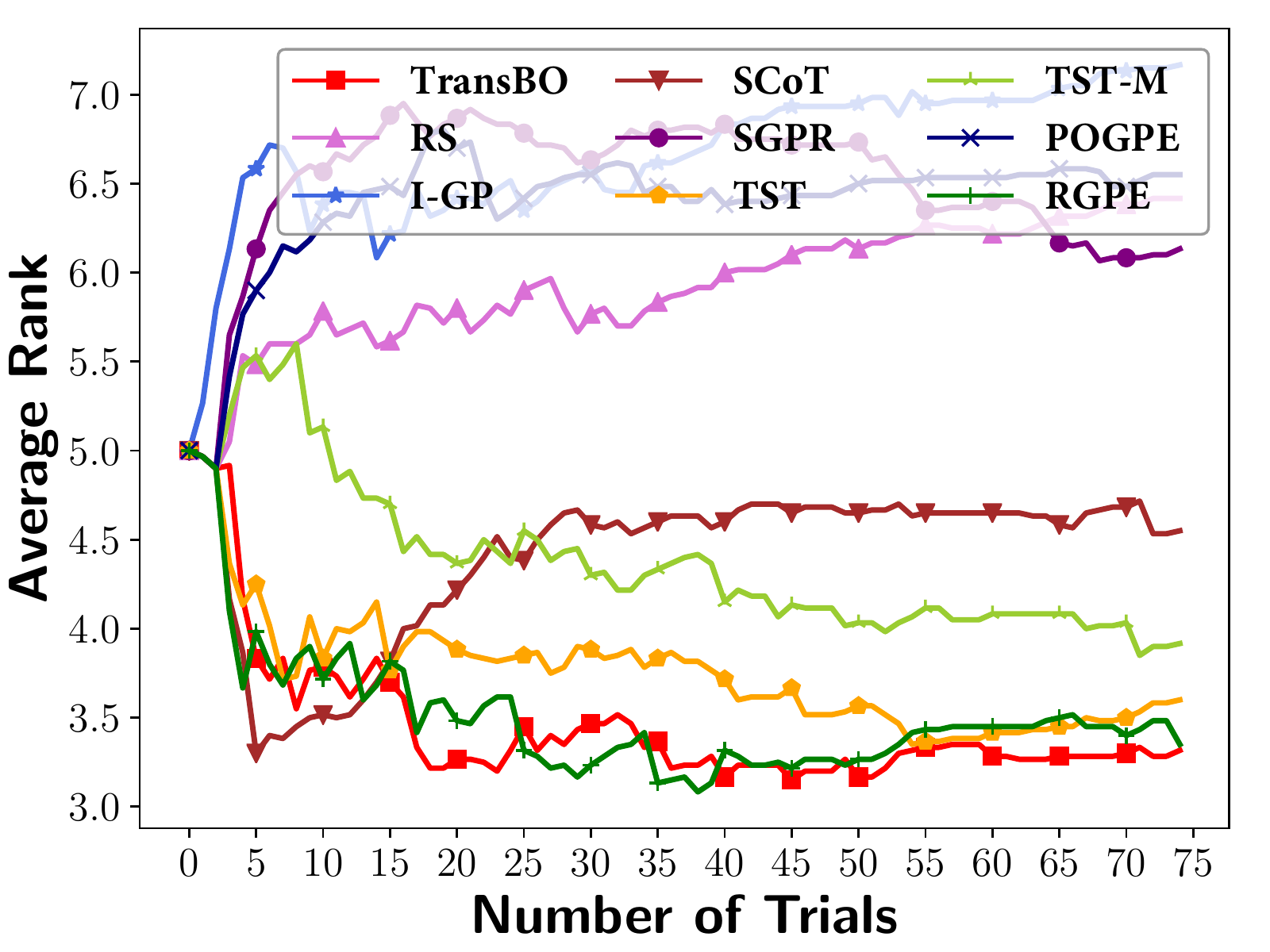}
			\label{offline_exp2_ext}
	}}
	\caption{Static TL results for four algorithms with $N_{task} = 5$ source tasks}

  \label{offline_exp2}
\end{figure*}

\subsection{Experimental Setup}
\para{\textbf{Baselines.}}
We compare \sys with eight baselines --
two non-transfer methods: (1) Random search~\cite{bergstra2012random}, (2) I-GP: independent Gaussian process-based surrogate fitted on the target task without using any source data, (3) SCoT~\cite{bardenet2013collaborative}: it models the relationship between datasets and hyperparamter performance by training a single surrogate on the scaled and merged observations from both source tasks and the target task, (4) SGPR: the core TL algorithm used in the well-known service --- Google Vizier~\cite{golovin2017google}, and four ensemble based TL methods: (5) POGPE~\cite{schilling2016scalable}, (6) TST~\cite{wistuba2016two}, (7) TST-M: a variant of TST using dataset meta-features~\cite{wistuba2016two}, and (8) RGPE~\cite{feurer2018scalable}.

\para{\textbf{Benchmark on 30 OpenML Datasets.}}
To evaluate the performance of \sys, we create and publish a large-scale benchmark. 
Four ML algorithms, including Random Forest, Extra Trees, Adaboost and LightGBM~\cite{ke2017lightgbm}, are tuned on 30 real-world datasets (tasks) from OpenML repository~\cite{10.1145/2641190.2641198}. 
The design of hyperparameter space and meta-feature for each dataset is adopted from the implementation in Auto-Sklearn~\cite{feurer2015efficient}. 
For each ML algorithm on each dataset, we sample 20k configurations from the hyperparameter space randomly and store the corresponding evaluation results.
It takes more than 200k CPU hours to collect these evaluation results. 
Note that, for reproducibility, we provide more details about this benchmark, including the datasets, the hyperparameter space of ML algorithms, etc., in Appendix~\ref{a.1}.

\para{\textbf{AutoML HPO Tasks.}}
To evaluate the performance of each method, the experiments are performed in a leave-one-out fashion. 
Each method optimizes the hyperparameters of a specific task over 20k configurations while treating the remaining tasks as the source tasks. 
In each source task, only $N_S$ instances (here $N_S=50$) are used to extract knowledge from this task in order to test the efficiency of TL~\cite{wistuba2016two, feurer2018scalable}.

We include the following three kinds of tasks:

(a)~\emph{\bf Static TL Setting.} This experiment is performed in a leave-one-out fashion, i.e., we optimize the hyperparameters of the target task while treating the remaining tasks as the source tasks. 

(b)~\emph{\bf Dynamic TL Setting.} It simulates the real-world HPO scenarios, in which 30 tasks (datasets) arrive sequentially; when the $i$-th task appears, the former $i-1$ tasks are treated as the source tasks.

(c)~\emph{\bf Neural Architecture Search (NAS).} It transfers tuning knowledge from conducting NAS on CIFAR-10 and CIFAR-100 to accelerate NAS on ImageNet16-120 based on NAS-Bench201~\cite{dong2019bench}.

In addition, following~\cite{feurer2018scalable}, all the compared methods are initialized with three randomly selected configurations, after which they proceed sequentially with a total of $N_{T}$ evaluations (trials). 
To avoid the effect of randomness, each method is repeated 30 times, and the averaged performance metrics are reported.

\para{Evaluation Metric.}
Comparing each method in terms of classification error is questionable because the classification error is not commensurable across datasets. 
Following the previous works~\cite{bardenet2013collaborative,wistuba2016two,feurer2018scalable}, we adopt the metrics as follows:

\emph{Average Rank.} For each target task, we rank all compared methods based on the performance of the best configuration they have found so far. 
Furthermore, ties are being solved by giving the average rank. 
For example, if one method observes the lowest validation error of 0.2, another two methods find 0.3, and the last method finds only 0.45, we would rank the methods with $1$, $\frac{2+3}{2}$, $\frac{2+3}{2}$, $4$.

\emph{Average Distance to Minimum.} The average distance to the global minimum after $t$ trials is defined as:
\begin{equation}
\small
ADTM(\mathcal{X}_t) = \frac{1}{K}\sum_{i \in [1:K]}\frac{min_{\bm{x} \in \mathcal{X}_t} y^{i}_{\bm{x}} - y^{i}_{min}}{y^i_{max} - y^{i}_{min}},
\end{equation}
where $y^i_{min}$ and $y^i_{max}$ are the best and worst performance value on the $i$-th task, $K$ is the number of tasks, i.e., $K=30$, $y_{\bm{x}}^i$ corresponds to the performance of configuration $\bm{x}$ in the $i$-th task, and $\mathcal{X}_t$ is the set of hyperparameter configurations that have been evaluated in the previous $t$ trials. 
The relative distances over all considered tasks are averaged to obtain the final ADTM value.

\para{Implementations \& Parameters.}
\sys implements the Gaussian process using SMAC3\footnote{https://github.com/automl/SMAC3}~\cite{hutter2011sequential,Lindauer2021SMAC3AV},
which can support a complex hyperparameter space, including numerical, categorical, and conditional hyperparameters, and the kernel hyperparameters in GP are inferred by maximizing the marginal likelihood. 
The two optimization problems in \sys are solved by using SQP methods provided in SciPy~\footnote{https://docs.scipy.org/doc/scipy/reference/optimize.minimize-slsqp.html}~\cite{2020SciPy-NMeth}.
In the BO module, the popular EI acquisition function is used.
As for the parameters in each baseline, the bandwidth $\rho$ in TST~\cite{wistuba2016two} is set to 0.3 for all experiments; in RGPE, we sample 100 times ($S=100$) to calculate the weight for each base surrogate; 
in SGPR~\cite{golovin2017google}, the parameter $\alpha$, which determines the relative importance of standard deviations of past tasks and the current task, is set to 0.95 (Check Appendix~\ref{reproduction} for reproduction details).

\begin{figure*}[htb]
	\centering
	\subfigure[Adaboost (Average Rank)]{
		\scalebox{0.23}[0.23]{
			\includegraphics[width=1\linewidth]{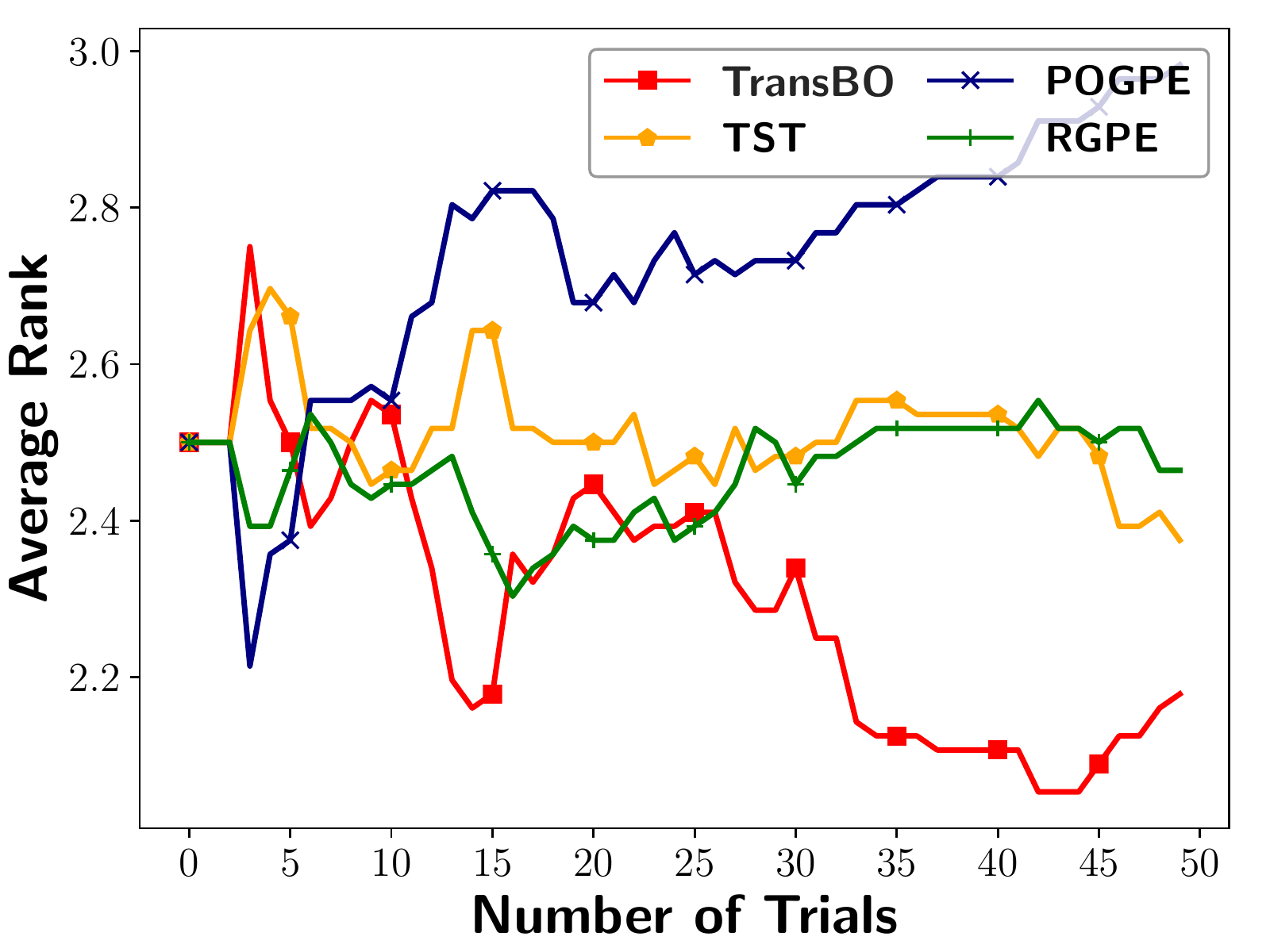}
	}}
	\subfigure[Adaboost (ADTM)]{
		\scalebox{0.23}[0.23]{
			\includegraphics[width=1\linewidth]{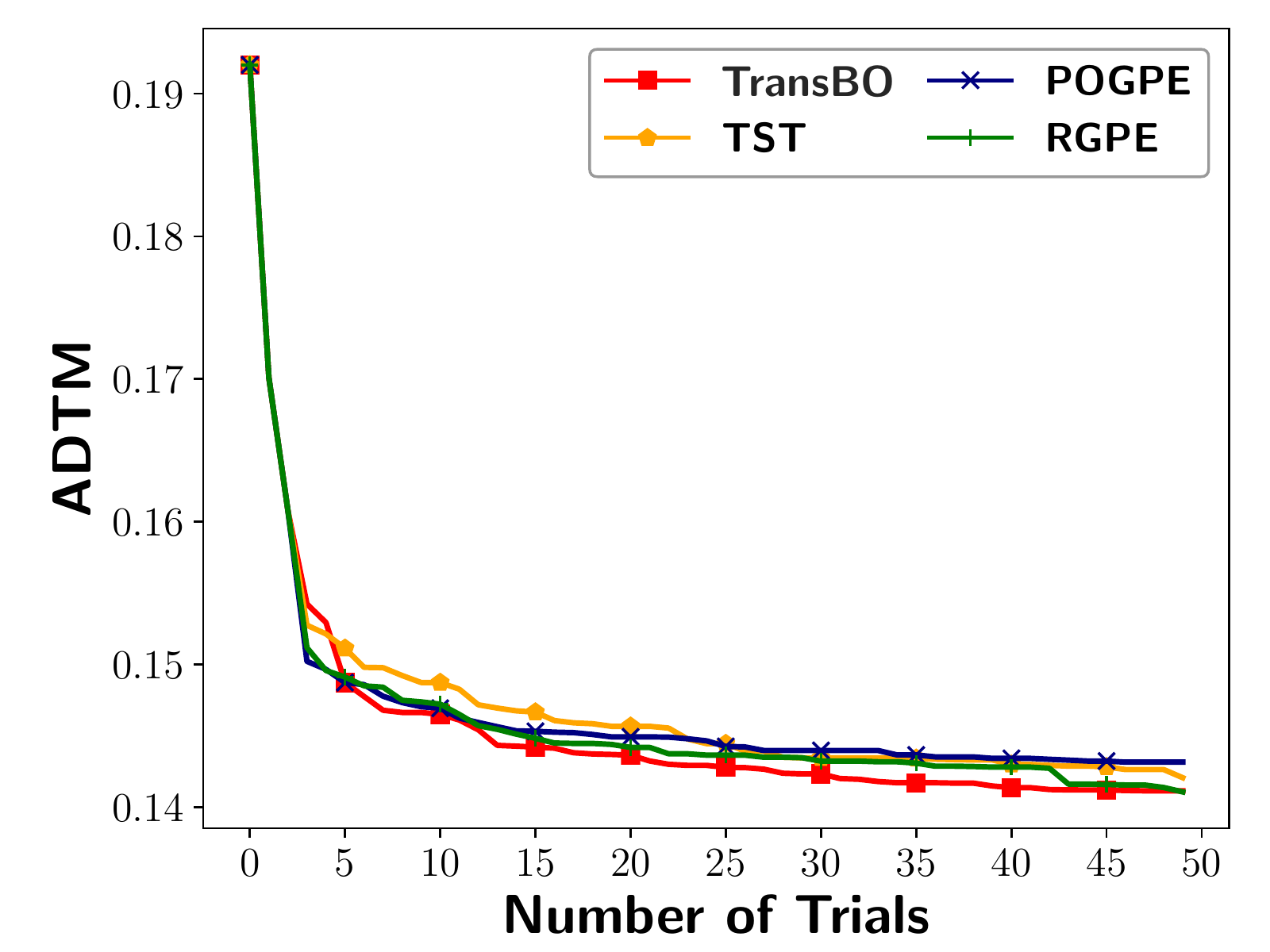}
	}}
	\subfigure[LightGBM (Average Rank)]{
		\scalebox{0.23}[0.23]{
			\includegraphics[width=1\linewidth]{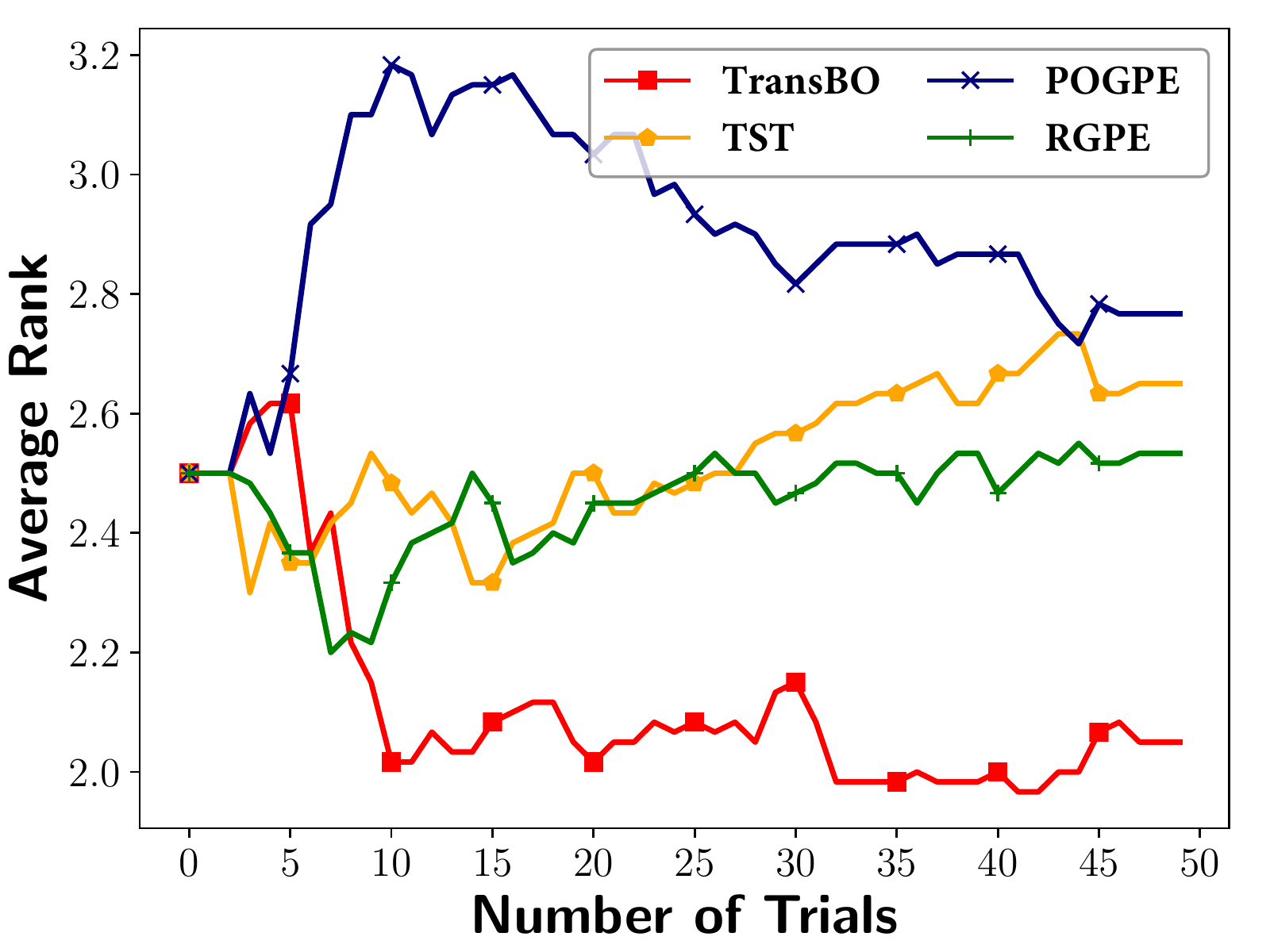}
	}}
    \subfigure[LightGBM (ADTM)]{
		\scalebox{0.23}[0.23]{
			\includegraphics[width=1\linewidth]{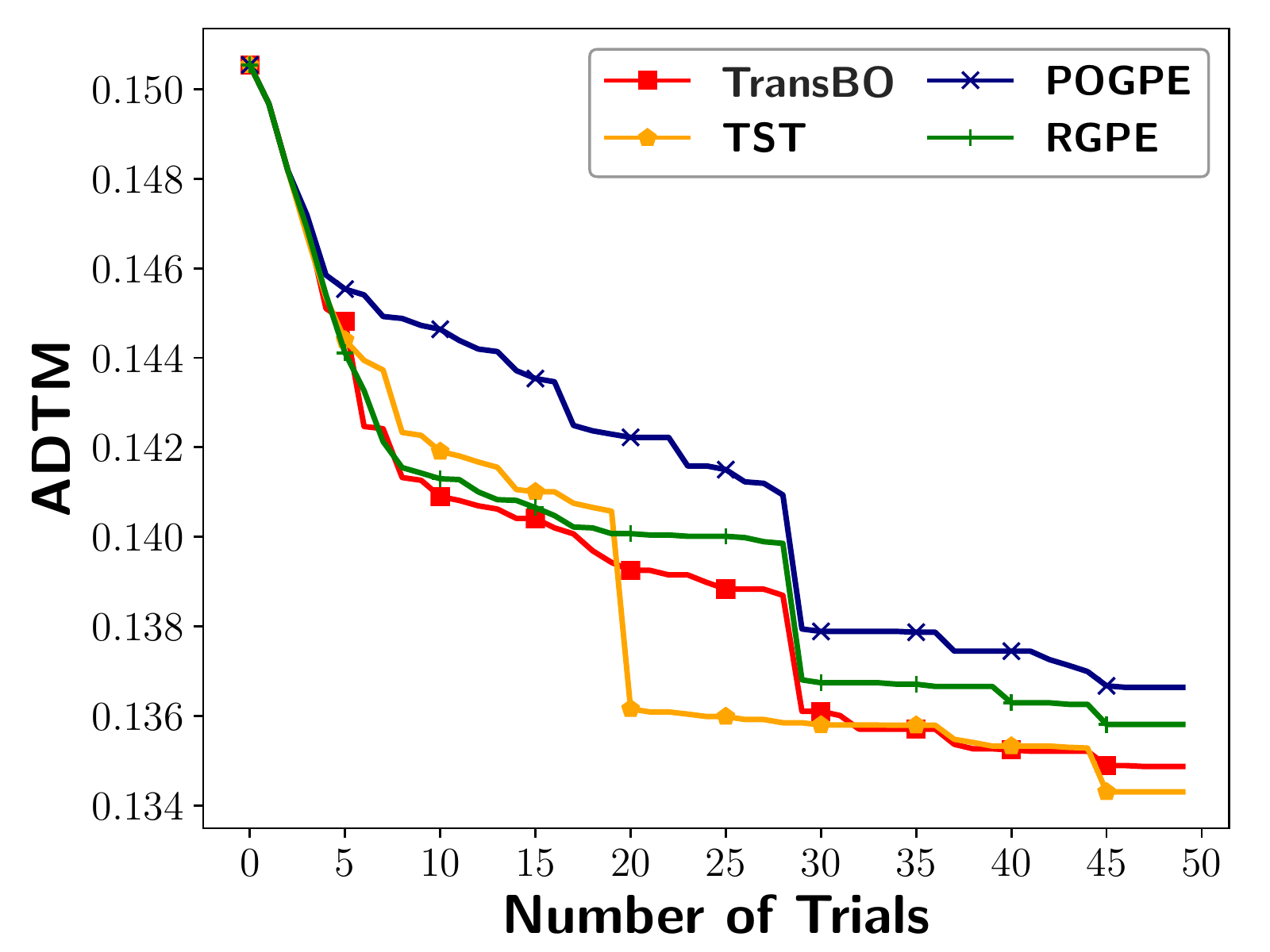}
	}}
	\caption{Results on source knowledge learning.}
  \label{source_ext_exp3}
\end{figure*}

\begin{figure*}[htb]
	\centering
	\subfigure[Random Forest (\emph{cpu\_small})]{
		\scalebox{0.23}[0.23]{
			\includegraphics[width=1\linewidth]{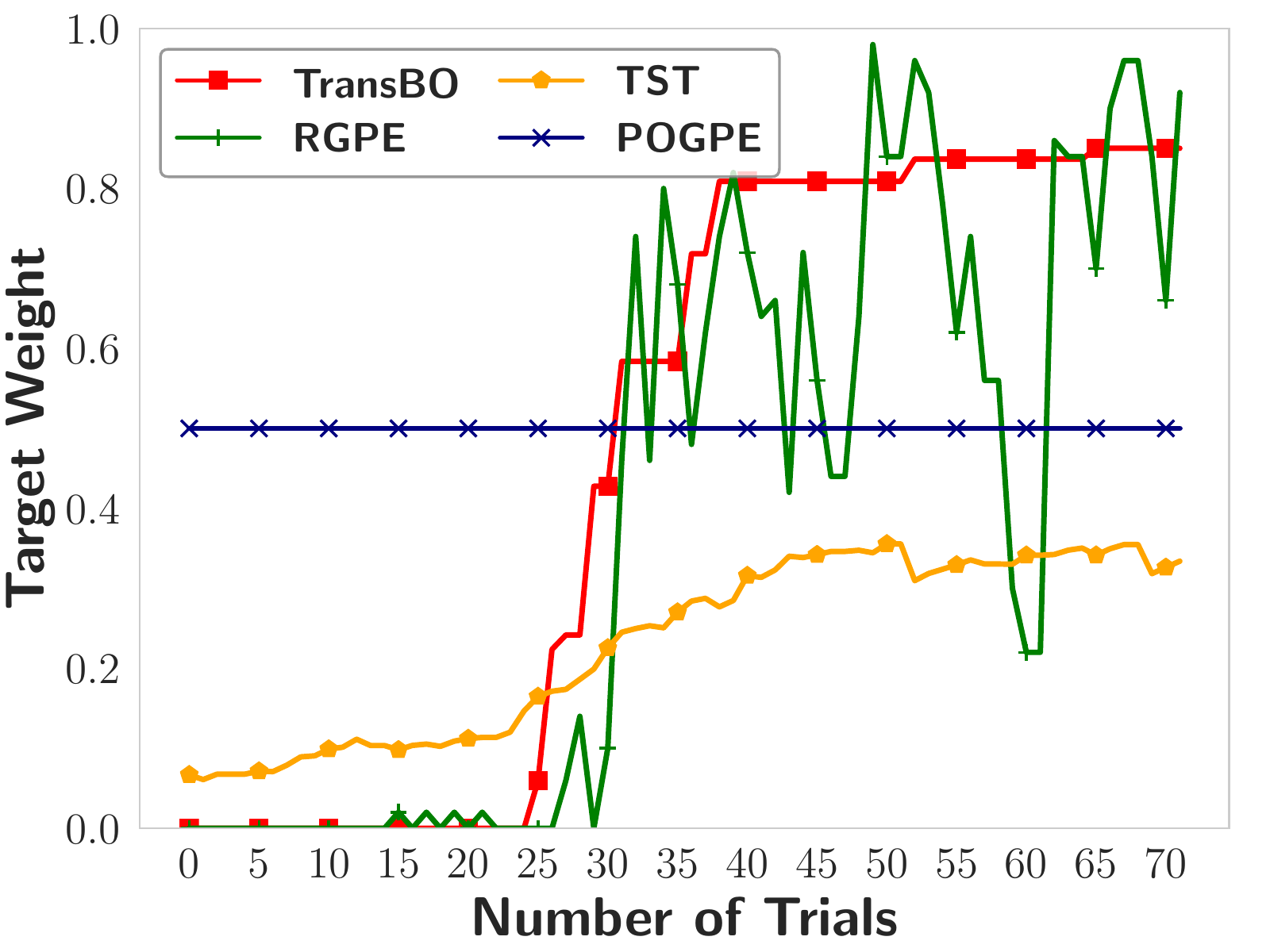}
			\label{weight_rf}
	}}
	\subfigure[Adaboost (\emph{hypothyroid(2)})]{
		\scalebox{0.23}[0.23]{
			\includegraphics[width=1\linewidth]{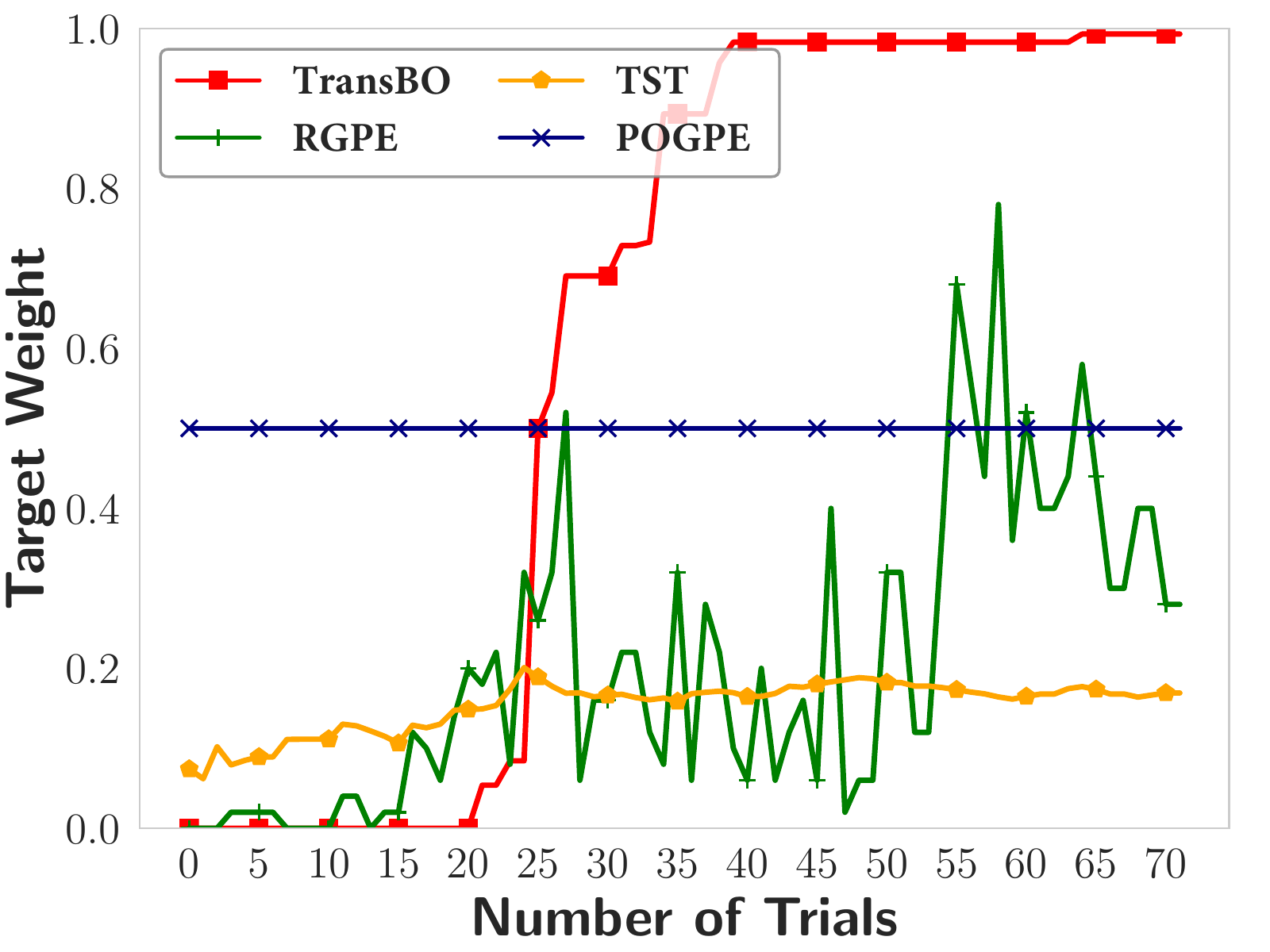}
			\label{weight_ada}
	}}
	\subfigure[Scalability on Random Forest]{
		\scalebox{0.23}[0.23]{
			\includegraphics[width=1\linewidth]{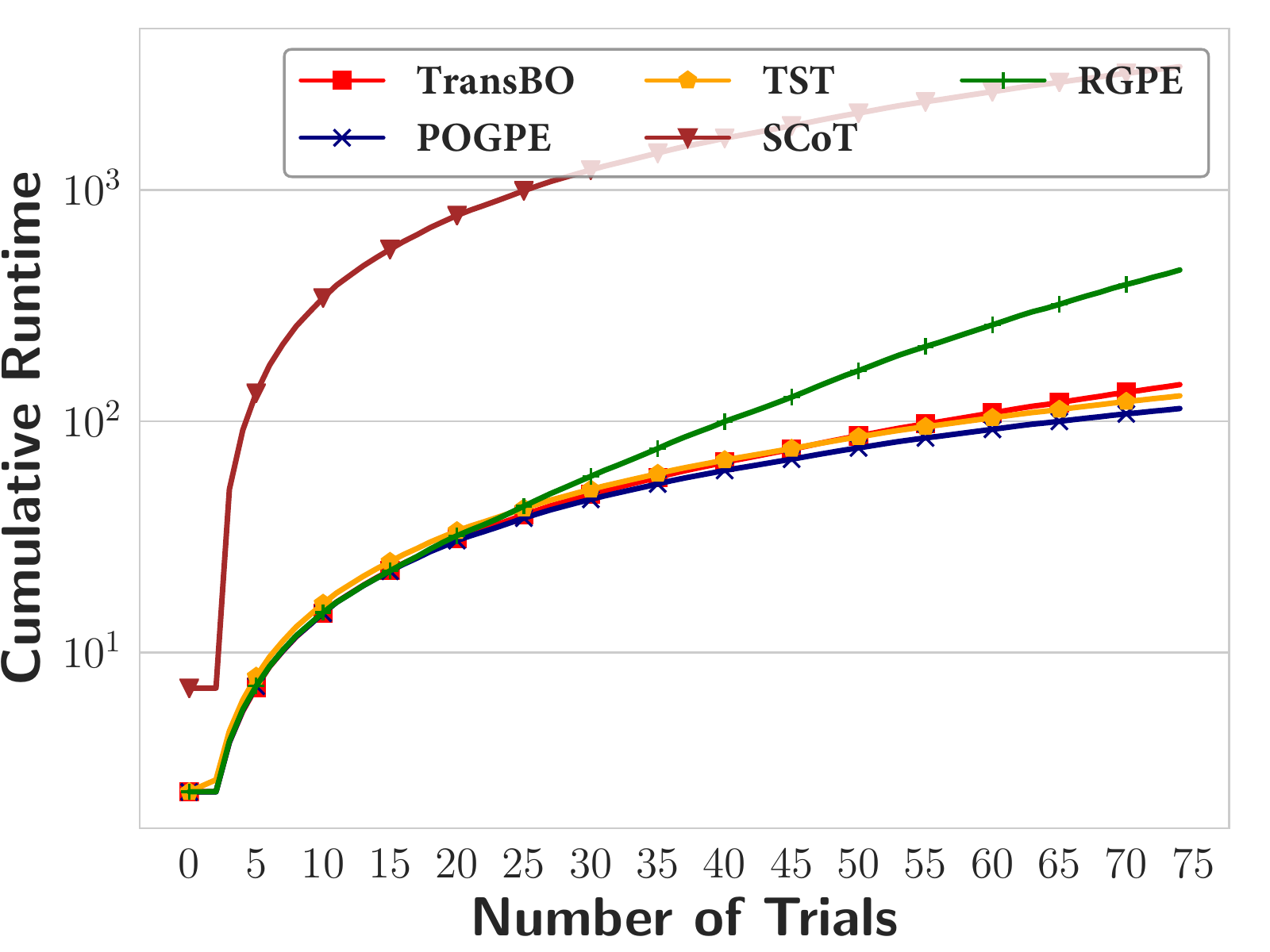}
			\label{runtime}
	}}
	\caption{Target weight and scalability analysis.}
  \label{algo_exp}
\end{figure*}

\subsection{Comprehensive Experiments in Two TL Settings}
\para{\bf Static TL Setting.} To demonstrate the efficiency and effectiveness of transfer learning in the static scenario, we compare \sys with the baselines on four benchmarks (i.e., Random Forest, LightGBM, Adaboost, and Extra Trees). 
Concretely, each task is selected as the target task in turn, and the remaining tasks are the source tasks;
then we can measure the performance of each baseline based on the results when tuning the hyperparameters of the target task.
Furthermore, we use 29 and 5 source tasks respectively to evaluate the ability of each method when given a different amount of source knowledge in terms of the number of source tasks $N_{task}$. 
Note that, for each target task, the maximum number of trials is 75. 
Figure~\ref{offline_exp1} and Figure~\ref{offline_exp2} show the experiment results on four benchmarks with 29 and 5 source tasks respectively, using average rank; more results on ADTM can be found in Appendix~\ref{a.3}.

First, we can observe that the average rank of \sys in Figure~\ref{offline_exp1} and Figure~\ref{offline_exp2} decreases sharply in the initial 20 trials.
Compared with other TL methods, it shows that \sys can extract and utilize the auxiliary source knowledge efficiently and effectively.
Remarkably, \sys exhibits a strong stability from two perspectives: 1) \sys is stable on different benchmarks; and 2) it still performs well when given a different number of source tasks, e.g., in Figure~\ref{offline_exp1} $N_{task}=29$, and $N_{task}=5$ in Figure~\ref{offline_exp2}.
RGPE is one of the most competitive baselines, and we take it as an example.
RGPE achieves comparable or similar performance with \sys in Figure~\ref{offline_exp1_lgb} and Figure~\ref{offline_exp1_adb} where $N_{task} = 29$. 
However, in Figure~\ref{offline_exp2_lgb} and Figure~\ref{offline_exp2_adb} RGPE exhibits a larger fluctuation over the trials compared with \sys when $N_{task} = 5$.
Unlike the baselines, \sys extracts the source knowledge in a principled way, and the empirical results show it performs well in most circumstances, thus demonstrating its superior efficiency and effectiveness.

\begin{table}[htb]
\centering
\caption{Dynamic TL results for Tuning four ML algorithms.}
\small
\resizebox{1\columnwidth}{!}{
  \begin{tabular}{lcccccccc}
    \toprule
\multirow{2}*{Method}  & \multicolumn{2}{c}{Adaboost} & \multicolumn{2}{c}{Random Forest} & \multicolumn{2}{c}{Extra Trees} & \multicolumn{2}{c}{LightGBM} \\
& 1st & 2nd & 1st & 2nd & 1st & 2nd & 1st & 2nd \\
\hline
POGPE & 0 & 2 & 0 & 1 & 0 & 2 & 1 & 2 \\
\hline 
TST & 8 & 12 & 9 & 9 & 7 & 12 & 10 & 9 \\
\hline 
RGPE & 8 & 5 & 6 & 14 & 10 & 9 & 9 & 10 \\
\hline
\textbf{\sys} & \textbf{14} & 11 & \textbf{15} & 6 & \textbf{14} & 7 & \textbf{12} & 10 \\
\bottomrule
  \end{tabular}
}
\label{online-table}
\end{table}

\para{Dynamic TL Setting.}
To simulate the real-world transfer learning scenario, we perform the dynamic experiment on different benchmarks. 
In this experiment, 30 tasks arrive sequentially; when the $i$-th task arrives, the previous $i$-1 tasks are used as the source tasks. 
The maximum number of trials for each task is 50, and we compare \sys with TST, RGPE, and POGPE based on the best-observed performance on each task. Table~\ref{online-table} reports the number of tasks on which each TL method gets the highest and second-highest performance. 
Note that the sum of each column may be more than 30 since some of the TL methods are tied for first or second place.

As shown in Table~\ref{online-table}, \sys achieves the largest number of top1 and top2 online performance among the compared methods. 
Take Adaboost as an example, \sys gets 25 top2 results among 30 tasks, while this number is 13 for RGPE.
RGPE gets a similar performance with TST on Lightgbm and Extra Trees, but its performance decreases on Adaboost. Thus, RGPE is not stable in this scenario.
Compared with the baselines, \sys could achieve more stable and satisfactory performance in the dynamic setting.

\subsection{Applying \sys to NAS}
\label{sec:apply_nas}
\begin{figure}[htb]
	\centering
	\scalebox{0.7}{
	  \includegraphics[width=1\linewidth]{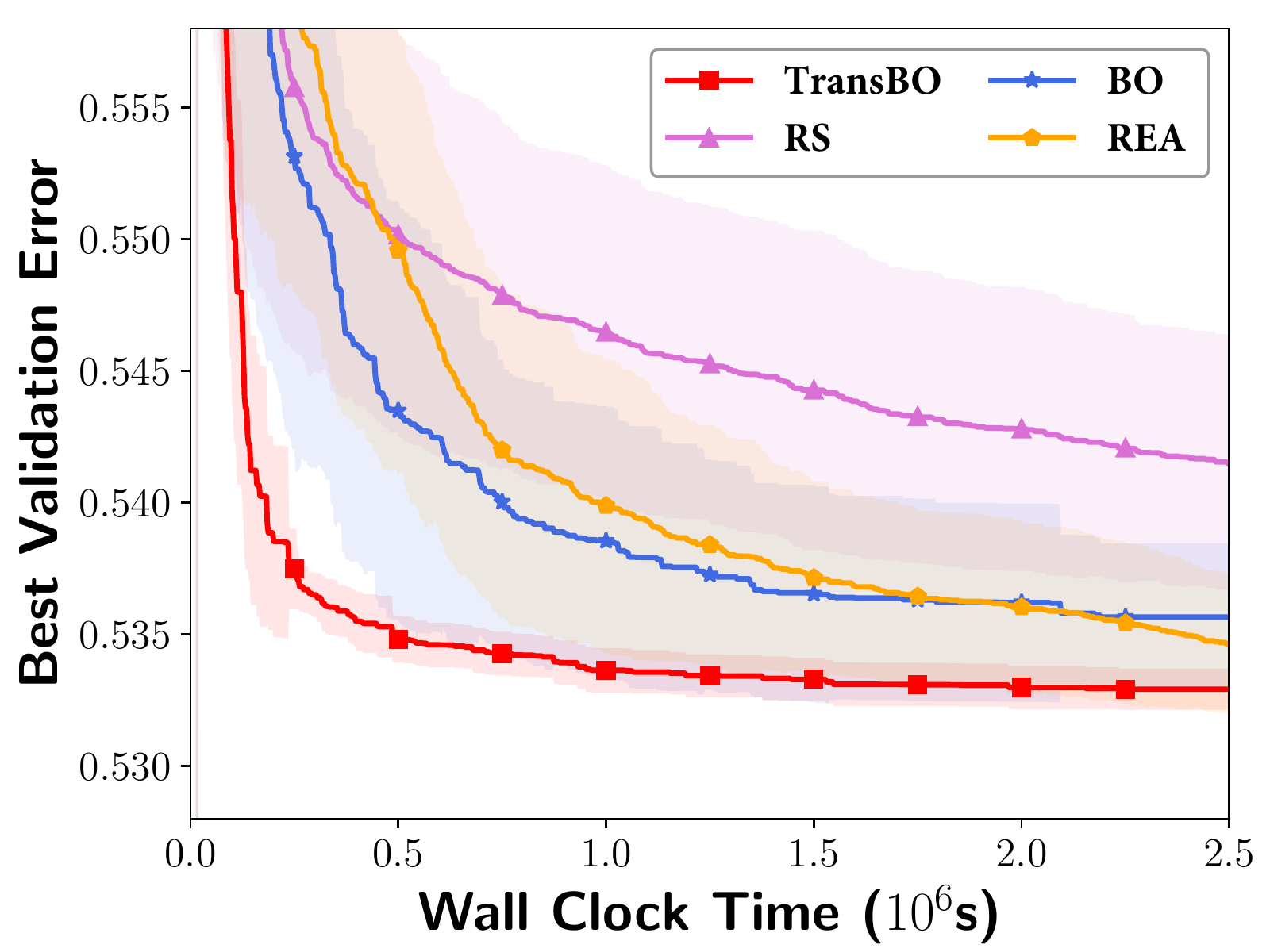}
	}
	\caption{Results on optimizing NAS on NASBench201.}
	\label{fig:nas}
\end{figure}

To investigate the universality of \sys in conducting Neural Architecture Search (NAS), here we use \sys to extract and integrate the optimization knowledge from NAS tasks on CIFAR-10 and CIFAR-100 (with 50 trials each) to accelerate the NAS task on ImageNet with NAS-Bench201~\cite{dong2019bench}. From Figure~\ref{fig:nas}, we have that \sys could achieve more than 5x speedups over the state-of-the-art NAS methods -- Bayesian Optimization (BO) and Regularized Evolution Algorithm (REA)~\cite{real2019regularized}.
Therefore, \sys can also be applied to the NAS tasks.

\subsection{Ablation Studies}
\label{sec:abla}
\para{Source Knowledge Learning.}
This experiment is designed to evaluate the performance of source surrogate $M^{S}$ learned in Phase 1. 
$M^{S}$ corresponds to the source knowledge extracted from the source tasks. 
In this setting, the source surrogate is used to guide the optimization of hyperparameters instead of the final TL surrogate $M^{TL}$. 
The quality of source knowledge learned by each TL method thus can be measured by the performance of $M^{S}$. 
Figure~\ref{source_ext_exp3} shows the results of \sys and three one-phase framework based methods: POGPE, TST, and RGPE on two benchmarks --- Adaboost and LightGBM. 
We can observe that the proposed \sys outperforms the other three baselines on both two metrics: average rank and ADTM.
According to some heuristics, these baselines calculate the weights in $M^{S}$ independently. 
Instead, by solving the constrained optimization problem, \sys can learn the optimal weights in $M^{s}$ in a joint and principled manner. 
More results on the other two benchmarks can be found in Appendix~\ref{a.3}.

\para{Target Weight Analysis.}
Here we compare the target weight obtained in POGPE, RGPE, TST, and \sys.
Figure~\ref{weight_rf} and ~\ref{weight_ada} illustrate the trend of target weight on two benchmarks: Random Forest and Adaboost. 
The target weight in POGPE is fixed to a constant - 0.5, regardless of the increasing number of trials; TST's remains low even when the target observations are sufficient; RGPE's shows a trend of fluctuation because the sampling-based ranking loss is not stable. 
\sys's keeps increasing with the number of trials, which matches the intuition that the importance of the target surrogate should be low when target observations are insufficient and gradually increase as target observations grow.

\para{Scalability Analysis.}
In the static TL setting, we include different number of source tasks when conducting transfer learning (See Figures~\ref{offline_exp1} and~\ref{offline_exp2}, where $N_{task}$ = 5 and $N_{task}$ = 29); the stable and effective results show the scalability in terms of the number of source tasks.
We further investigate the optimization overhead of suggesting a new configuration, and measure the runtime of the baselines - POGPE, RGPE, TST, SCoT, and \sys on Random Forest with 75 trials. 
To investigate the scalability of \sys, we measure the runtime of the competitive TL methods: POGPE, RGPE, TST, SCoT, and \sys. 
Each method is tested on Random Forest with 75 trials, and we repeat each method 20 times.
Figure \ref{runtime} shows the experiment results, where the y-axis is the mean cumulative runtime in seconds on a log scale.
We do not take the evaluation time of each configuration into account, and only compare the optimization overhead of suggesting a new configuration. 
ScoT's runtime increases rapidly among the compared methods as it has the $O(k^3n^3)$ complexity.
Since both the two-phase and one-phase framework-based methods own the $O(n^3)$ complexity, it takes nearly the same optimization overhead for TST, POGPE, and \sys to suggest a configuration in the first 75 trials.
Although RGPE also has the $O(n^3)$ complexity, it depends on a sampling strategy to compute the surrogate weight, which introduces extra overhead to configuration suggestion.
Instead, \sys exhibits a similar scalability result like POGPE, which incorporates no optimization overhead due to the constant weights. 
This shows that \sys scales well in both the number of trials and tasks.

\section{Conclusion}
In this paper, we introduced \sys, a novel two-phase transfer learning (TL) method for hyperparameter optimization (HPO), which can leverage the auxiliary knowledge from previous tasks to accelerate the HPO process of the current task effectively.
This framework can extract and aggregate the source and target knowledge jointly and adaptively.
In addition, we published a large-scale TL benchmark for HPO with up to 1.8 million model evaluations; the extensive experiments, including static and dynamic transfer learning settings and neural architecture search, demonstrate the superiority of \sys over the state-of-the-art methods.

\begin{acks}
This work was supported by the National Natural Science Foundation of China (No.61832001), Beijing Academy of Artificial Intelligence (BAAI), PKU-Tencent Joint Research Lab. Bin Cui is the corresponding author.
\end{acks}

\bibliographystyle{ACM-Reference-Format}
\bibliography{reference}


\begin{thebibliography}{60}


\ifx \showCODEN    \undefined \def \showCODEN     #1{\unskip}     \fi
\ifx \showDOI      \undefined \def \showDOI       #1{#1}\fi
\ifx \showISBNx    \undefined \def \showISBNx     #1{\unskip}     \fi
\ifx \showISBNxiii \undefined \def \showISBNxiii  #1{\unskip}     \fi
\ifx \showISSN     \undefined \def \showISSN      #1{\unskip}     \fi
\ifx \showLCCN     \undefined \def \showLCCN      #1{\unskip}     \fi
\ifx \shownote     \undefined \def \shownote      #1{#1}          \fi
\ifx \showarticletitle \undefined \def \showarticletitle #1{#1}   \fi
\ifx \showURL      \undefined \def \showURL       {\relax}        \fi
\providecommand\bibfield[2]{#2}
\providecommand\bibinfo[2]{#2}
\providecommand\natexlab[1]{#1}
\providecommand\showeprint[2][]{arXiv:#2}

\bibitem[Bardenet et~al\mbox{.}(2013)]%
        {bardenet2013collaborative}
\bibfield{author}{\bibinfo{person}{R{\'e}mi Bardenet},
  \bibinfo{person}{M{\'a}ty{\'a}s Brendel}, \bibinfo{person}{Bal{\'a}zs
  K{\'e}gl}, {and} \bibinfo{person}{Michele Sebag}.}
  \bibinfo{year}{2013}\natexlab{}.
\newblock \showarticletitle{Collaborative hyperparameter tuning}. In
  \bibinfo{booktitle}{\emph{ICML}}. \bibinfo{pages}{199--207}.
\newblock


\bibitem[Bennett et~al\mbox{.}(2008)]%
        {bennett2008bilevel}
\bibfield{author}{\bibinfo{person}{Kristin~P Bennett}, \bibinfo{person}{Gautam
  Kunapuli}, \bibinfo{person}{Jing Hu}, {and} \bibinfo{person}{Jong-Shi Pang}.}
  \bibinfo{year}{2008}\natexlab{}.
\newblock \showarticletitle{Bilevel optimization and machine learning}. In
  \bibinfo{booktitle}{\emph{IEEE World Congress on Computational
  Intelligence}}. Springer, \bibinfo{pages}{25--47}.
\newblock


\bibitem[Bergstra and Bengio(2012)]%
        {bergstra2012random}
\bibfield{author}{\bibinfo{person}{James Bergstra} {and}
  \bibinfo{person}{Yoshua Bengio}.} \bibinfo{year}{2012}\natexlab{}.
\newblock \showarticletitle{Random search for hyper-parameter optimization}.
\newblock \bibinfo{journal}{\emph{Journal of Machine Learning Research}}
  \bibinfo{volume}{13}, \bibinfo{number}{Feb} (\bibinfo{year}{2012}),
  \bibinfo{pages}{281--305}.
\newblock


\bibitem[Bergstra et~al\mbox{.}(2011)]%
        {bergstra2011algorithms}
\bibfield{author}{\bibinfo{person}{James~S Bergstra}, \bibinfo{person}{R{\'e}mi
  Bardenet}, \bibinfo{person}{Yoshua Bengio}, {and} \bibinfo{person}{Bal{\'a}zs
  K{\'e}gl}.} \bibinfo{year}{2011}\natexlab{}.
\newblock \showarticletitle{Algorithms for hyper-parameter optimization}. In
  \bibinfo{booktitle}{\emph{Advances in neural information processing
  systems}}. \bibinfo{pages}{2546--2554}.
\newblock


\bibitem[Bischl et~al\mbox{.}(2021)]%
        {bischl2021hyperparameter}
\bibfield{author}{\bibinfo{person}{Bernd Bischl}, \bibinfo{person}{Martin
  Binder}, \bibinfo{person}{Michel Lang}, \bibinfo{person}{Tobias Pielok},
  \bibinfo{person}{Jakob Richter}, \bibinfo{person}{Stefan Coors},
  \bibinfo{person}{Janek Thomas}, \bibinfo{person}{Theresa Ullmann},
  \bibinfo{person}{Marc Becker}, \bibinfo{person}{Anne-Laure Boulesteix},
  {et~al\mbox{.}}} \bibinfo{year}{2021}\natexlab{}.
\newblock \showarticletitle{Hyperparameter optimization: Foundations,
  algorithms, best practices and open challenges}.
\newblock \bibinfo{journal}{\emph{arXiv preprint arXiv:2107.05847}}
  (\bibinfo{year}{2021}).
\newblock


\bibitem[Devlin et~al\mbox{.}(2018)]%
        {devlin2018bert}
\bibfield{author}{\bibinfo{person}{Jacob Devlin}, \bibinfo{person}{Ming-Wei
  Chang}, \bibinfo{person}{Kenton Lee}, {and} \bibinfo{person}{Kristina
  Toutanova}.} \bibinfo{year}{2018}\natexlab{}.
\newblock \showarticletitle{Bert: Pre-training of deep bidirectional
  transformers for language understanding}.
\newblock \bibinfo{journal}{\emph{arXiv preprint arXiv:1810.04805}}
  (\bibinfo{year}{2018}).
\newblock


\bibitem[Dong and Yang(2019)]%
        {dong2019bench}
\bibfield{author}{\bibinfo{person}{Xuanyi Dong} {and} \bibinfo{person}{Yi
  Yang}.} \bibinfo{year}{2019}\natexlab{}.
\newblock \showarticletitle{NAS-Bench-201: Extending the Scope of Reproducible
  Neural Architecture Search}. In \bibinfo{booktitle}{\emph{International
  Conference on Learning Representations}}.
\newblock


\bibitem[Dudziak et~al\mbox{.}(2020)]%
        {dudziak2020brp}
\bibfield{author}{\bibinfo{person}{Lukasz Dudziak}, \bibinfo{person}{Thomas
  Chau}, \bibinfo{person}{Mohamed Abdelfattah}, \bibinfo{person}{Royson Lee},
  \bibinfo{person}{Hyeji Kim}, {and} \bibinfo{person}{Nicholas Lane}.}
  \bibinfo{year}{2020}\natexlab{}.
\newblock \showarticletitle{BRP-NAS: Prediction-based NAS using GCNs}.
\newblock \bibinfo{journal}{\emph{Advances in Neural Information Processing
  Systems}}  \bibinfo{volume}{33} (\bibinfo{year}{2020}).
\newblock


\bibitem[{Falkner} et~al\mbox{.}(2018)]%
        {falkner2018bohb}
\bibfield{author}{\bibinfo{person}{Stefan {Falkner}}, \bibinfo{person}{Aaron
  {Klein}}, {and} \bibinfo{person}{Frank {Hutter}}.}
  \bibinfo{year}{2018}\natexlab{}.
\newblock \showarticletitle{BOHB: Robust and Efficient Hyperparameter
  Optimization at Scale.}. In \bibinfo{booktitle}{\emph{ICML}}.
  \bibinfo{pages}{1436--1445}.
\newblock


\bibitem[Feurer et~al\mbox{.}(2015)]%
        {feurer2015efficient}
\bibfield{author}{\bibinfo{person}{Matthias Feurer}, \bibinfo{person}{Aaron
  Klein}, \bibinfo{person}{Katharina Eggensperger}, \bibinfo{person}{Jost
  Springenberg}, \bibinfo{person}{Manuel Blum}, {and} \bibinfo{person}{Frank
  Hutter}.} \bibinfo{year}{2015}\natexlab{}.
\newblock \showarticletitle{Efficient and robust automated machine learning}.
  In \bibinfo{booktitle}{\emph{Advances in neural information processing
  systems}}. \bibinfo{pages}{2962--2970}.
\newblock


\bibitem[Feurer et~al\mbox{.}(2018)]%
        {feurer2018scalable}
\bibfield{author}{\bibinfo{person}{Matthias Feurer}, \bibinfo{person}{Benjamin
  Letham}, {and} \bibinfo{person}{Eytan Bakshy}.}
  \bibinfo{year}{2018}\natexlab{}.
\newblock \showarticletitle{Scalable meta-learning for bayesian optimization
  using ranking-weighted gaussian process ensembles}. In
  \bibinfo{booktitle}{\emph{AutoML Workshop at ICML}}.
\newblock


\bibitem[Foster et~al\mbox{.}(2019)]%
        {NEURIPS2019_d55cbf21}
\bibfield{author}{\bibinfo{person}{Adam Foster}, \bibinfo{person}{Martin
  Jankowiak}, \bibinfo{person}{Elias Bingham}, \bibinfo{person}{Paul Horsfall},
  \bibinfo{person}{Yee~Whye Teh}, \bibinfo{person}{Thomas Rainforth}, {and}
  \bibinfo{person}{Noah Goodman}.} \bibinfo{year}{2019}\natexlab{}.
\newblock \showarticletitle{Variational Bayesian optimal experimental design}.
\newblock \bibinfo{journal}{\emph{Advances in Neural Information Processing
  Systems}}  \bibinfo{volume}{32} (\bibinfo{year}{2019}).
\newblock


\bibitem[Golovin et~al\mbox{.}(2017)]%
        {golovin2017google}
\bibfield{author}{\bibinfo{person}{Daniel Golovin}, \bibinfo{person}{Benjamin
  Solnik}, \bibinfo{person}{Subhodeep Moitra}, \bibinfo{person}{Greg
  Kochanski}, \bibinfo{person}{John Karro}, {and} \bibinfo{person}{D Sculley}.}
  \bibinfo{year}{2017}\natexlab{}.
\newblock \showarticletitle{Google vizier: A service for black-box
  optimization}. In \bibinfo{booktitle}{\emph{Proceedings of the 23rd ACM
  SIGKDD International Conference on Knowledge Discovery and Data Mining}}.
  ACM, \bibinfo{pages}{1487--1495}.
\newblock


\bibitem[Goodfellow et~al\mbox{.}(2016)]%
        {goodfellow2016deep}
\bibfield{author}{\bibinfo{person}{Ian Goodfellow}, \bibinfo{person}{Yoshua
  Bengio}, {and} \bibinfo{person}{Aaron Courville}.}
  \bibinfo{year}{2016}\natexlab{}.
\newblock \bibinfo{booktitle}{\emph{Deep learning}}.
\newblock \bibinfo{publisher}{MIT press}.
\newblock


\bibitem[He et~al\mbox{.}(2016)]%
        {he2016deep}
\bibfield{author}{\bibinfo{person}{Kaiming He}, \bibinfo{person}{Xiangyu
  Zhang}, \bibinfo{person}{Shaoqing Ren}, {and} \bibinfo{person}{Jian Sun}.}
  \bibinfo{year}{2016}\natexlab{}.
\newblock \showarticletitle{Deep residual learning for image recognition}. In
  \bibinfo{booktitle}{\emph{Proceedings of the IEEE conference on computer
  vision and pattern recognition}}. \bibinfo{pages}{770--778}.
\newblock


\bibitem[He et~al\mbox{.}(2017)]%
        {he2017neural}
\bibfield{author}{\bibinfo{person}{Xiangnan He}, \bibinfo{person}{Lizi Liao},
  \bibinfo{person}{Hanwang Zhang}, \bibinfo{person}{Liqiang Nie},
  \bibinfo{person}{Xia Hu}, {and} \bibinfo{person}{Tat-Seng Chua}.}
  \bibinfo{year}{2017}\natexlab{}.
\newblock \showarticletitle{Neural collaborative filtering}. In
  \bibinfo{booktitle}{\emph{Proceedings of the 26th international conference on
  world wide web}}. \bibinfo{pages}{173--182}.
\newblock


\bibitem[Henrique et~al\mbox{.}(2019)]%
        {henrique2019literature}
\bibfield{author}{\bibinfo{person}{Bruno~Miranda Henrique},
  \bibinfo{person}{Vinicius~Amorim Sobreiro}, {and} \bibinfo{person}{Herbert
  Kimura}.} \bibinfo{year}{2019}\natexlab{}.
\newblock \showarticletitle{Literature review: Machine learning techniques
  applied to financial market prediction}.
\newblock \bibinfo{journal}{\emph{Expert Systems with Applications}}
  \bibinfo{volume}{124} (\bibinfo{year}{2019}), \bibinfo{pages}{226--251}.
\newblock


\bibitem[Hinton et~al\mbox{.}(2012)]%
        {hinton2012deep}
\bibfield{author}{\bibinfo{person}{Geoffrey Hinton}, \bibinfo{person}{Li Deng},
  \bibinfo{person}{Dong Yu}, \bibinfo{person}{George~E Dahl},
  \bibinfo{person}{Abdel-rahman Mohamed}, \bibinfo{person}{Navdeep Jaitly},
  \bibinfo{person}{Andrew Senior}, \bibinfo{person}{Vincent Vanhoucke},
  \bibinfo{person}{Patrick Nguyen}, \bibinfo{person}{Tara~N Sainath},
  {et~al\mbox{.}}} \bibinfo{year}{2012}\natexlab{}.
\newblock \showarticletitle{Deep neural networks for acoustic modeling in
  speech recognition: The shared views of four research groups}.
\newblock \bibinfo{journal}{\emph{IEEE Signal processing magazine}}
  \bibinfo{volume}{29}, \bibinfo{number}{6} (\bibinfo{year}{2012}),
  \bibinfo{pages}{82--97}.
\newblock


\bibitem[Horv{\'a}th et~al\mbox{.}(2021)]%
        {horvath2021hyperparameter}
\bibfield{author}{\bibinfo{person}{Samuel Horv{\'a}th}, \bibinfo{person}{Aaron
  Klein}, \bibinfo{person}{Peter Richt{\'a}rik}, {and}
  \bibinfo{person}{C{\'e}dric Archambeau}.} \bibinfo{year}{2021}\natexlab{}.
\newblock \showarticletitle{Hyperparameter transfer learning with adaptive
  complexity}. In \bibinfo{booktitle}{\emph{International Conference on
  Artificial Intelligence and Statistics}}. PMLR, \bibinfo{pages}{1378--1386}.
\newblock


\bibitem[Hutter et~al\mbox{.}(2011)]%
        {hutter2011sequential}
\bibfield{author}{\bibinfo{person}{Frank Hutter}, \bibinfo{person}{Holger~H
  Hoos}, {and} \bibinfo{person}{Kevin Leyton-Brown}.}
  \bibinfo{year}{2011}\natexlab{}.
\newblock \showarticletitle{Sequential model-based optimization for general
  algorithm configuration}. In \bibinfo{booktitle}{\emph{International
  Conference on Learning and Intelligent Optimization}}. Springer,
  \bibinfo{pages}{507--523}.
\newblock


\bibitem[Jones et~al\mbox{.}(1998)]%
        {jones1998efficient}
\bibfield{author}{\bibinfo{person}{Donald~R Jones}, \bibinfo{person}{Matthias
  Schonlau}, {and} \bibinfo{person}{William~J Welch}.}
  \bibinfo{year}{1998}\natexlab{}.
\newblock \showarticletitle{Efficient global optimization of expensive
  black-box functions}.
\newblock \bibinfo{journal}{\emph{Journal of Global optimization}}
  \bibinfo{volume}{13}, \bibinfo{number}{4} (\bibinfo{year}{1998}),
  \bibinfo{pages}{455--492}.
\newblock


\bibitem[Joy et~al\mbox{.}(2016)]%
        {joy2016flexible}
\bibfield{author}{\bibinfo{person}{Tinu~Theckel Joy}, \bibinfo{person}{Santu
  Rana}, \bibinfo{person}{Sunil~Kumar Gupta}, {and} \bibinfo{person}{Svetha
  Venkatesh}.} \bibinfo{year}{2016}\natexlab{}.
\newblock \showarticletitle{Flexible transfer learning framework for Bayesian
  optimisation}. In \bibinfo{booktitle}{\emph{Pacific-Asia Conference on
  Knowledge Discovery and Data Mining}}. Springer, \bibinfo{pages}{102--114}.
\newblock


\bibitem[Kandasamy et~al\mbox{.}(2017)]%
        {kandasamy2017multi}
\bibfield{author}{\bibinfo{person}{Kirthevasan Kandasamy},
  \bibinfo{person}{Gautam Dasarathy}, \bibinfo{person}{Jeff Schneider}, {and}
  \bibinfo{person}{Barnab{\'a}s P{\'o}czos}.} \bibinfo{year}{2017}\natexlab{}.
\newblock \showarticletitle{Multi-fidelity bayesian optimisation with
  continuous approximations}. In \bibinfo{booktitle}{\emph{ICML}}. PMLR,
  \bibinfo{pages}{1799--1808}.
\newblock


\bibitem[Ke et~al\mbox{.}(2017)]%
        {ke2017lightgbm}
\bibfield{author}{\bibinfo{person}{Guolin Ke}, \bibinfo{person}{Qi Meng},
  \bibinfo{person}{Thomas Finley}, \bibinfo{person}{Taifeng Wang},
  \bibinfo{person}{Wei Chen}, \bibinfo{person}{Weidong Ma},
  \bibinfo{person}{Qiwei Ye}, {and} \bibinfo{person}{Tie-Yan Liu}.}
  \bibinfo{year}{2017}\natexlab{}.
\newblock \showarticletitle{Lightgbm: A highly efficient gradient boosting
  decision tree}.
\newblock \bibinfo{journal}{\emph{Advances in neural information processing
  systems}}  \bibinfo{volume}{30} (\bibinfo{year}{2017}).
\newblock


\bibitem[Kim et~al\mbox{.}(2017)]%
        {kim2017learning}
\bibfield{author}{\bibinfo{person}{Jungtaek Kim}, \bibinfo{person}{Saehoon
  Kim}, {and} \bibinfo{person}{Seungjin Choi}.}
  \bibinfo{year}{2017}\natexlab{}.
\newblock \showarticletitle{Learning to Transfer Initializations for Bayesian
  Hyperparameter Optimization}.
\newblock \bibinfo{journal}{\emph{ArXiv}}  \bibinfo{volume}{abs/1710.06219}
  (\bibinfo{year}{2017}).
\newblock


\bibitem[Klein et~al\mbox{.}(2017a)]%
        {klein2017fast}
\bibfield{author}{\bibinfo{person}{Aaron Klein}, \bibinfo{person}{Stefan
  Falkner}, \bibinfo{person}{Simon Bartels}, \bibinfo{person}{Philipp Hennig},
  \bibinfo{person}{Frank Hutter}, {et~al\mbox{.}}}
  \bibinfo{year}{2017}\natexlab{a}.
\newblock \showarticletitle{Fast Bayesian hyperparameter optimization on large
  datasets}.
\newblock \bibinfo{journal}{\emph{Electronic Journal of Statistics}}
  \bibinfo{volume}{11}, \bibinfo{number}{2} (\bibinfo{year}{2017}),
  \bibinfo{pages}{4945--4968}.
\newblock


\bibitem[Klein et~al\mbox{.}(2017b)]%
        {klein2016learning}
\bibfield{author}{\bibinfo{person}{Aaron Klein}, \bibinfo{person}{S. Falkner},
  \bibinfo{person}{Jost~Tobias Springenberg}, {and} \bibinfo{person}{F.
  Hutter}.} \bibinfo{year}{2017}\natexlab{b}.
\newblock \showarticletitle{Learning Curve Prediction with Bayesian Neural
  Networks}. In \bibinfo{booktitle}{\emph{ICLR}}.
\newblock


\bibitem[Kraft(1994)]%
        {10.1145/192115.192124}
\bibfield{author}{\bibinfo{person}{Dieter Kraft}.}
  \bibinfo{year}{1994}\natexlab{}.
\newblock \showarticletitle{Algorithm 733: TOMP--Fortran modules for optimal
  control calculations}.
\newblock \bibinfo{journal}{\emph{ACM Transactions on Mathematical Software
  (TOMS)}} \bibinfo{volume}{20}, \bibinfo{number}{3} (\bibinfo{year}{1994}),
  \bibinfo{pages}{262--281}.
\newblock


\bibitem[Li et~al\mbox{.}(2020)]%
        {li2020efficient}
\bibfield{author}{\bibinfo{person}{Yang Li}, \bibinfo{person}{Jiawei Jiang},
  \bibinfo{person}{Jinyang Gao}, \bibinfo{person}{Yingxia Shao},
  \bibinfo{person}{Ce Zhang}, {and} \bibinfo{person}{Bin Cui}.}
  \bibinfo{year}{2020}\natexlab{}.
\newblock \showarticletitle{Efficient automatic CASH via rising bandits}. In
  \bibinfo{booktitle}{\emph{Proceedings of the AAAI Conference on Artificial
  Intelligence}}, Vol.~\bibinfo{volume}{34}. \bibinfo{pages}{4763--4771}.
\newblock


\bibitem[Li et~al\mbox{.}(2022)]%
        {li2022hyper}
\bibfield{author}{\bibinfo{person}{Yang Li}, \bibinfo{person}{Yu Shen},
  \bibinfo{person}{Huaijun Jiang}, \bibinfo{person}{Wentao Zhang},
  \bibinfo{person}{Jixiang Li}, \bibinfo{person}{Ji Liu}, \bibinfo{person}{Ce
  Zhang}, {and} \bibinfo{person}{Bin Cui}.} \bibinfo{year}{2022}\natexlab{}.
\newblock \showarticletitle{Hyper-Tune: Towards Efficient Hyper-parameter
  Tuning at Scale}.
\newblock \bibinfo{journal}{\emph{Proceedings of the VLDB Endowment}}
  \bibinfo{volume}{15} (\bibinfo{year}{2022}).
\newblock


\bibitem[Li et~al\mbox{.}(2021a)]%
        {li-mfeshb}
\bibfield{author}{\bibinfo{person}{Yang Li}, \bibinfo{person}{Yu Shen},
  \bibinfo{person}{Jiawei Jiang}, \bibinfo{person}{Jinyang Gao},
  \bibinfo{person}{Ce Zhang}, {and} \bibinfo{person}{Bin Cui}.}
  \bibinfo{year}{2021}\natexlab{a}.
\newblock \showarticletitle{MFES-HB: Efficient Hyperband with Multi-Fidelity
  Quality Measurements}. In \bibinfo{booktitle}{\emph{Proceedings of the AAAI
  Conference on Artificial Intelligence}}, Vol.~\bibinfo{volume}{35}. AAAI
  Press, \bibinfo{pages}{8491--8500}.
\newblock


\bibitem[Li et~al\mbox{.}(2021b)]%
        {openbox}
\bibfield{author}{\bibinfo{person}{Yang Li}, \bibinfo{person}{Yu Shen},
  \bibinfo{person}{Wentao Zhang}, \bibinfo{person}{Yuanwei Chen},
  \bibinfo{person}{Huaijun Jiang}, \bibinfo{person}{Mingchao Liu},
  \bibinfo{person}{Jiawei Jiang}, \bibinfo{person}{Jinyang Gao},
  \bibinfo{person}{Wentao Wu}, \bibinfo{person}{Zhi Yang}, {et~al\mbox{.}}}
  \bibinfo{year}{2021}\natexlab{b}.
\newblock \showarticletitle{Openbox: A generalized black-box optimization
  service}. In \bibinfo{booktitle}{\emph{Proceedings of the 27th ACM SIGKDD
  Conference on Knowledge Discovery \& Data Mining}}.
  \bibinfo{pages}{3209--3219}.
\newblock


\bibitem[Li et~al\mbox{.}(2021c)]%
        {li2021volcanoml}
\bibfield{author}{\bibinfo{person}{Yang Li}, \bibinfo{person}{Yu Shen},
  \bibinfo{person}{Wentao Zhang}, \bibinfo{person}{Jiawei Jiang},
  \bibinfo{person}{Bolin Ding}, \bibinfo{person}{Yaliang Li},
  \bibinfo{person}{Jingren Zhou}, \bibinfo{person}{Zhi Yang},
  \bibinfo{person}{Wentao Wu}, \bibinfo{person}{Ce Zhang}, {et~al\mbox{.}}}
  \bibinfo{year}{2021}\natexlab{c}.
\newblock \showarticletitle{VolcanoML: speeding up end-to-end AutoML via
  scalable search space decomposition}.
\newblock \bibinfo{journal}{\emph{Proceedings of the VLDB Endowment}}
  \bibinfo{volume}{14} (\bibinfo{year}{2021}).
\newblock


\bibitem[Lindauer and Hutter(2018)]%
        {lindauer2018warmstarting}
\bibfield{author}{\bibinfo{person}{Marius Lindauer} {and}
  \bibinfo{person}{Frank Hutter}.} \bibinfo{year}{2018}\natexlab{}.
\newblock \showarticletitle{Warmstarting of model-based algorithm
  configuration}. In \bibinfo{booktitle}{\emph{Thirty-Second AAAI Conference on
  Artificial Intelligence}}.
\newblock


\bibitem[Lindauer et~al\mbox{.}(2021)]%
        {Lindauer2021SMAC3AV}
\bibfield{author}{\bibinfo{person}{Marius~Thomas Lindauer},
  \bibinfo{person}{Katharina Eggensperger}, \bibinfo{person}{Matthias Feurer},
  \bibinfo{person}{Andr'e Biedenkapp}, \bibinfo{person}{Difan Deng},
  \bibinfo{person}{Caroline Benjamins}, \bibinfo{person}{Ren{\'e} Sass}, {and}
  \bibinfo{person}{Frank Hutter}.} \bibinfo{year}{2021}\natexlab{}.
\newblock \showarticletitle{SMAC3: A Versatile Bayesian Optimization Package
  for Hyperparameter Optimization}.
\newblock \bibinfo{journal}{\emph{ArXiv}}  \bibinfo{volume}{abs/2109.09831}
  (\bibinfo{year}{2021}).
\newblock


\bibitem[{Pan} and {Yang}(2010)]%
        {pan2010a}
\bibfield{author}{\bibinfo{person}{Sinno~Jialin {Pan}} {and}
  \bibinfo{person}{Qiang {Yang}}.} \bibinfo{year}{2010}\natexlab{}.
\newblock \showarticletitle{A Survey on Transfer Learning}.
\newblock \bibinfo{journal}{\emph{IEEE Transactions on Knowledge and Data
  Engineering}} \bibinfo{volume}{22}, \bibinfo{number}{10}
  (\bibinfo{year}{2010}), \bibinfo{pages}{1345--1359}.
\newblock


\bibitem[Pan et~al\mbox{.}(2010)]%
        {pan2010survey}
\bibfield{author}{\bibinfo{person}{Sinno~Jialin Pan}, \bibinfo{person}{Qiang
  Yang}, {et~al\mbox{.}}} \bibinfo{year}{2010}\natexlab{}.
\newblock \showarticletitle{A survey on transfer learning}.
\newblock \bibinfo{journal}{\emph{IEEE Transactions on knowledge and data
  engineering}} \bibinfo{volume}{22}, \bibinfo{number}{10}
  (\bibinfo{year}{2010}), \bibinfo{pages}{1345--1359}.
\newblock


\bibitem[Pardoe and Stone(2010)]%
        {pardoe2010boosting}
\bibfield{author}{\bibinfo{person}{David Pardoe} {and} \bibinfo{person}{Peter
  Stone}.} \bibinfo{year}{2010}\natexlab{}.
\newblock \showarticletitle{Boosting for regression transfer}. In
  \bibinfo{booktitle}{\emph{ICML}}. \bibinfo{pages}{863--870}.
\newblock


\bibitem[Perrone et~al\mbox{.}(2018)]%
        {perrone2018scalable}
\bibfield{author}{\bibinfo{person}{Valerio Perrone}, \bibinfo{person}{Rodolphe
  Jenatton}, \bibinfo{person}{Matthias~W Seeger}, {and}
  \bibinfo{person}{C{\'e}dric Archambeau}.} \bibinfo{year}{2018}\natexlab{}.
\newblock \showarticletitle{Scalable hyperparameter transfer learning}.
\newblock \bibinfo{journal}{\emph{Advances in neural information processing
  systems}}  \bibinfo{volume}{31} (\bibinfo{year}{2018}).
\newblock


\bibitem[Perrone et~al\mbox{.}(2019)]%
        {NIPS2019_9438}
\bibfield{author}{\bibinfo{person}{Valerio Perrone}, \bibinfo{person}{Huibin
  Shen}, \bibinfo{person}{Matthias~W Seeger}, \bibinfo{person}{Cedric
  Archambeau}, {and} \bibinfo{person}{Rodolphe Jenatton}.}
  \bibinfo{year}{2019}\natexlab{}.
\newblock \showarticletitle{Learning search spaces for bayesian optimization:
  Another view of hyperparameter transfer learning}.
\newblock \bibinfo{journal}{\emph{Advances in Neural Information Processing
  Systems}}  \bibinfo{volume}{32} (\bibinfo{year}{2019}).
\newblock


\bibitem[Poloczek et~al\mbox{.}(2017)]%
        {poloczek2017multi}
\bibfield{author}{\bibinfo{person}{Matthias Poloczek}, \bibinfo{person}{Jialei
  Wang}, {and} \bibinfo{person}{Peter Frazier}.}
  \bibinfo{year}{2017}\natexlab{}.
\newblock \showarticletitle{Multi-information source optimization}. In
  \bibinfo{booktitle}{\emph{Advances in Neural Information Processing
  Systems}}. \bibinfo{pages}{4288--4298}.
\newblock


\bibitem[Rasmussen(2004)]%
        {rasmussen2004gaussian}
\bibfield{author}{\bibinfo{person}{Carl~Edward Rasmussen}.}
  \bibinfo{year}{2004}\natexlab{}.
\newblock \showarticletitle{Gaussian processes in machine learning}.
\newblock In \bibinfo{booktitle}{\emph{Advanced lectures on machine learning}}.
  \bibinfo{publisher}{Springer}, \bibinfo{pages}{63--71}.
\newblock


\bibitem[Real et~al\mbox{.}(2019)]%
        {real2019regularized}
\bibfield{author}{\bibinfo{person}{Esteban Real}, \bibinfo{person}{Alok
  Aggarwal}, \bibinfo{person}{Yanping Huang}, {and} \bibinfo{person}{Quoc~V
  Le}.} \bibinfo{year}{2019}\natexlab{}.
\newblock \showarticletitle{Regularized evolution for image classifier
  architecture search}. In \bibinfo{booktitle}{\emph{Proceedings of the AAAI
  conference on artificial intelligence}}, Vol.~\bibinfo{volume}{33}.
  \bibinfo{pages}{4780--4789}.
\newblock


\bibitem[Salinas et~al\mbox{.}(2020)]%
        {salinas2020quantile}
\bibfield{author}{\bibinfo{person}{David Salinas}, \bibinfo{person}{Huibin
  Shen}, {and} \bibinfo{person}{Valerio Perrone}.}
  \bibinfo{year}{2020}\natexlab{}.
\newblock \showarticletitle{A quantile-based approach for hyperparameter
  transfer learning}. In \bibinfo{booktitle}{\emph{ICML}}. PMLR,
  \bibinfo{pages}{8438--8448}.
\newblock


\bibitem[Schilling et~al\mbox{.}(2015)]%
        {schilling2015hyperparameter}
\bibfield{author}{\bibinfo{person}{Nicolas Schilling}, \bibinfo{person}{Martin
  Wistuba}, \bibinfo{person}{Lucas Drumond}, {and} \bibinfo{person}{Lars
  Schmidt-Thieme}.} \bibinfo{year}{2015}\natexlab{}.
\newblock \showarticletitle{Hyperparameter optimization with factorized
  multilayer perceptrons}. In \bibinfo{booktitle}{\emph{ECML PKDD}}.
  \bibinfo{pages}{87--103}.
\newblock


\bibitem[Schilling et~al\mbox{.}(2016)]%
        {schilling2016scalable}
\bibfield{author}{\bibinfo{person}{Nicolas Schilling}, \bibinfo{person}{Martin
  Wistuba}, {and} \bibinfo{person}{Lars Schmidt-Thieme}.}
  \bibinfo{year}{2016}\natexlab{}.
\newblock \showarticletitle{Scalable hyperparameter optimization with products
  of gaussian process experts}. In \bibinfo{booktitle}{\emph{ECML PKDD}}.
  Springer, \bibinfo{pages}{33--48}.
\newblock


\bibitem[Sinha et~al\mbox{.}(2017)]%
        {sinha2017review}
\bibfield{author}{\bibinfo{person}{Ankur Sinha}, \bibinfo{person}{Pekka Malo},
  {and} \bibinfo{person}{Kalyanmoy Deb}.} \bibinfo{year}{2017}\natexlab{}.
\newblock \showarticletitle{A review on bilevel optimization: from classical to
  evolutionary approaches and applications}.
\newblock \bibinfo{journal}{\emph{IEEE Transactions on Evolutionary
  Computation}} \bibinfo{volume}{22}, \bibinfo{number}{2}
  (\bibinfo{year}{2017}), \bibinfo{pages}{276--295}.
\newblock


\bibitem[Snoek et~al\mbox{.}(2012)]%
        {snoek2012practical}
\bibfield{author}{\bibinfo{person}{Jasper Snoek}, \bibinfo{person}{Hugo
  Larochelle}, {and} \bibinfo{person}{Ryan~P Adams}.}
  \bibinfo{year}{2012}\natexlab{}.
\newblock \showarticletitle{Practical bayesian optimization of machine learning
  algorithms}. In \bibinfo{booktitle}{\emph{Advances in neural information
  processing systems}}. \bibinfo{pages}{2951--2959}.
\newblock


\bibitem[Snoek et~al\mbox{.}(2015)]%
        {snoek2015scalable}
\bibfield{author}{\bibinfo{person}{Jasper Snoek}, \bibinfo{person}{Oren
  Rippel}, \bibinfo{person}{Kevin Swersky}, \bibinfo{person}{Ryan Kiros},
  \bibinfo{person}{Nadathur Satish}, \bibinfo{person}{Narayanan Sundaram},
  \bibinfo{person}{Mostofa Patwary}, \bibinfo{person}{Mr Prabhat}, {and}
  \bibinfo{person}{Ryan Adams}.} \bibinfo{year}{2015}\natexlab{}.
\newblock \showarticletitle{Scalable bayesian optimization using deep neural
  networks}. In \bibinfo{booktitle}{\emph{ICML}}. PMLR,
  \bibinfo{pages}{2171--2180}.
\newblock


\bibitem[Springenberg et~al\mbox{.}(2016)]%
        {springenberg2016bayesian}
\bibfield{author}{\bibinfo{person}{Jost~Tobias Springenberg},
  \bibinfo{person}{Aaron Klein}, \bibinfo{person}{Stefan Falkner}, {and}
  \bibinfo{person}{Frank Hutter}.} \bibinfo{year}{2016}\natexlab{}.
\newblock \showarticletitle{Bayesian optimization with robust Bayesian neural
  networks}.
\newblock \bibinfo{journal}{\emph{Advances in neural information processing
  systems}}  \bibinfo{volume}{29} (\bibinfo{year}{2016}).
\newblock


\bibitem[Swersky et~al\mbox{.}(2013)]%
        {swersky2013multi}
\bibfield{author}{\bibinfo{person}{Kevin Swersky}, \bibinfo{person}{Jasper
  Snoek}, {and} \bibinfo{person}{Ryan~P Adams}.}
  \bibinfo{year}{2013}\natexlab{}.
\newblock \showarticletitle{Multi-task bayesian optimization}.
\newblock \bibinfo{journal}{\emph{Advances in neural information processing
  systems}}  \bibinfo{volume}{26} (\bibinfo{year}{2013}).
\newblock


\bibitem[Swersky et~al\mbox{.}(2014)]%
        {swersky2014freeze}
\bibfield{author}{\bibinfo{person}{Kevin Swersky}, \bibinfo{person}{Jasper
  Snoek}, {and} \bibinfo{person}{Ryan~Prescott Adams}.}
  \bibinfo{year}{2014}\natexlab{}.
\newblock \showarticletitle{Freeze-thaw Bayesian optimization}.
\newblock \bibinfo{journal}{\emph{arXiv preprint arXiv:1406.3896}}
  (\bibinfo{year}{2014}).
\newblock


\bibitem[Vanschoren et~al\mbox{.}(2014)]%
        {10.1145/2641190.2641198}
\bibfield{author}{\bibinfo{person}{Joaquin Vanschoren}, \bibinfo{person}{Jan~N
  Van~Rijn}, \bibinfo{person}{Bernd Bischl}, {and} \bibinfo{person}{Luis
  Torgo}.} \bibinfo{year}{2014}\natexlab{}.
\newblock \showarticletitle{OpenML: networked science in machine learning}.
\newblock \bibinfo{journal}{\emph{ACM SIGKDD Explorations Newsletter}}
  \bibinfo{volume}{15}, \bibinfo{number}{2} (\bibinfo{year}{2014}),
  \bibinfo{pages}{49--60}.
\newblock


\bibitem[Virtanen et~al\mbox{.}(2020)]%
        {2020SciPy-NMeth}
\bibfield{author}{\bibinfo{person}{Pauli Virtanen}, \bibinfo{person}{Ralf
  Gommers}, \bibinfo{person}{Travis~E Oliphant}, \bibinfo{person}{Matt
  Haberland}, \bibinfo{person}{Tyler Reddy}, \bibinfo{person}{David
  Cournapeau}, \bibinfo{person}{Evgeni Burovski}, \bibinfo{person}{Pearu
  Peterson}, \bibinfo{person}{Warren Weckesser}, \bibinfo{person}{Jonathan
  Bright}, {et~al\mbox{.}}} \bibinfo{year}{2020}\natexlab{}.
\newblock \showarticletitle{SciPy 1.0: fundamental algorithms for scientific
  computing in Python}.
\newblock \bibinfo{journal}{\emph{Nature methods}} \bibinfo{volume}{17},
  \bibinfo{number}{3} (\bibinfo{year}{2020}), \bibinfo{pages}{261--272}.
\newblock


\bibitem[Wei et~al\mbox{.}(2021)]%
        {wei2021meta}
\bibfield{author}{\bibinfo{person}{Ying Wei}, \bibinfo{person}{Peilin Zhao},
  {and} \bibinfo{person}{Junzhou Huang}.} \bibinfo{year}{2021}\natexlab{}.
\newblock \showarticletitle{Meta-learning Hyperparameter Performance Prediction
  with Neural Processes}. In \bibinfo{booktitle}{\emph{ICML}}. PMLR,
  \bibinfo{pages}{11058--11067}.
\newblock


\bibitem[Wistuba et~al\mbox{.}(2015a)]%
        {wistuba2015hyperparameter}
\bibfield{author}{\bibinfo{person}{Martin Wistuba}, \bibinfo{person}{Nicolas
  Schilling}, {and} \bibinfo{person}{Lars Schmidt-Thieme}.}
  \bibinfo{year}{2015}\natexlab{a}.
\newblock \showarticletitle{Hyperparameter search space pruning--a new
  component for sequential model-based hyperparameter optimization}. In
  \bibinfo{booktitle}{\emph{ECML PKDD}}. Springer, \bibinfo{pages}{104--119}.
\newblock


\bibitem[Wistuba et~al\mbox{.}(2015b)]%
        {wistuba2015sequential}
\bibfield{author}{\bibinfo{person}{Martin Wistuba}, \bibinfo{person}{Nicolas
  Schilling}, {and} \bibinfo{person}{Lars Schmidt-Thieme}.}
  \bibinfo{year}{2015}\natexlab{b}.
\newblock \showarticletitle{Sequential model-free hyperparameter tuning}. In
  \bibinfo{booktitle}{\emph{Data Mining (ICDM), 2015 IEEE International
  Conference on}}. IEEE, \bibinfo{pages}{1033--1038}.
\newblock


\bibitem[Wistuba et~al\mbox{.}(2016)]%
        {wistuba2016two}
\bibfield{author}{\bibinfo{person}{Martin Wistuba}, \bibinfo{person}{Nicolas
  Schilling}, {and} \bibinfo{person}{Lars Schmidt-Thieme}.}
  \bibinfo{year}{2016}\natexlab{}.
\newblock \showarticletitle{Two-stage transfer surrogate model for automatic
  hyperparameter optimization}. In \bibinfo{booktitle}{\emph{ECML PKDD}}.
  \bibinfo{pages}{199--214}.
\newblock


\bibitem[Yao et~al\mbox{.}(2018)]%
        {quanming2018taking}
\bibfield{author}{\bibinfo{person}{Quanming Yao}, \bibinfo{person}{Mengshuo
  Wang}, \bibinfo{person}{H. Escalante}, \bibinfo{person}{I. Guyon},
  \bibinfo{person}{Yi-Qi Hu}, \bibinfo{person}{Yu-Feng Li},
  \bibinfo{person}{Wei-Wei Tu}, \bibinfo{person}{Qiang Yang}, {and}
  \bibinfo{person}{Yang Yu}.} \bibinfo{year}{2018}\natexlab{}.
\newblock \showarticletitle{Taking Human out of Learning Applications: A Survey
  on Automated Machine Learning}.
\newblock \bibinfo{journal}{\emph{ArXiv}}  \bibinfo{volume}{abs/1810.13306}
  (\bibinfo{year}{2018}).
\newblock


\bibitem[Yogatama and Mann(2014)]%
        {yogatama2014efficient}
\bibfield{author}{\bibinfo{person}{Dani Yogatama} {and} \bibinfo{person}{Gideon
  Mann}.} \bibinfo{year}{2014}\natexlab{}.
\newblock \showarticletitle{Efficient transfer learning method for automatic
  hyperparameter tuning}. In \bibinfo{booktitle}{\emph{Artificial Intelligence
  and Statistics}}. \bibinfo{pages}{1077--1085}.
\newblock


\end{thebibliography}

\clearpage

\appendix
\section{Appendix}

\subsection{The Details of Benchmark}
\label{a.1}
As described in Section~\ref{sec:exp_sec}, we create a benchmark to evaluate the performance of TL methods. 
We choose four ML algorithms that are widely used in data analysis, including Random Forest, Extra Trees, Adaboost and Lightgbm. 
The implementation of each algorithm and the design of their hyperparameter space follows Auto-sklearn. 
For each algorithm, the range and default value of each hyperparameter are illustrated in Tables ~\ref{hp_adaboost}, \ref{hp_trees} and \ref{hp_lgb}. 
To collect sufficient source HPO data for transfer learning, we select 30 real-world datasets from OpenML repository, and evaluate the validation performance (i.e., the balanced accuracy) of 20k configurations for each benchmark, which are randomly sampled from the hyperparameter space. 
The datasets used in our benchmarks are of medium size, whose number of rows ranges from 2000 to 8192. For more details, see Table~\ref{cls_datasets}. 
The total number of model evaluations (observations) in our benchmarks reaches 1.8 million and it takes more than 100k CPU hours to evaluate all the configurations. 
For reproduction purposes, we also upload the benchmark data (e.g., evaluation results and the corresponding scripts) along with this submission.
The benchmark data (with size – 477.7Mb); due to the space limit (maximum 20Mb) on CMT3, we only upload a small subset of benchmark on one algorithm — LightGBM.
After the review process, we will make the complete benchmark publicly available (e.g., on Google Drive).

\begin{table}[h]
\centering
\small
\begin{tabular}{lccc}
    \toprule
    Hyperparameter & Range & Default \\
    \midrule
    n\_estimators & [50, 500] & 50 \\
    learning\_rate (log) & [0.01, 2] & 0.1 \\
    algorithm & \{SAMME.R, SAMME\} & SAMME.R \\
    max\_depth & [2, 8] & 3 \\
    \bottomrule
\end{tabular}
\caption {Hyperparameters of Adaboost.}
\label{hp_adaboost}
\end{table}

\begin{table}[h]
\centering
\small
\begin{tabular}{lccc}
    \toprule
    Hyperparameter & Range  & Default \\
    \midrule
    criterion &  \{gini, entropy\} & gini \\
    max\_features & [0, 1] & 0.5 \\
    min\_sample\_split & [2, 20] & 2 \\
    min\_sample\_leaf & [1, 20] & 1 \\
    bootstrap & \{True, False\} & True \\
    \bottomrule
\end{tabular}
\caption {Hyperparameters of Random Forest and Extra Trees.}
\label{hp_trees}
\end{table}

\begin{table}[h]
\centering
\small
\begin{tabular}{lccc}
    \toprule
    Hyperparameter & Range  & Default \\
    \midrule
    n\_estimators & [100, 1000] & 500 \\
    num\_leaves & [31, 2047] & 127 \\
    learning\_rate (log) & [0.001, 0.3] & 0.1 \\
    min\_child\_samples & [5, 30] & 20 \\
    subsample & [0.7, 1] & 1 \\
    colsample\_bytree & [0.7, 1] & 1 \\
    \bottomrule
\end{tabular}
\caption {Hyperparameters of LightGBM.}
\label{hp_lgb}
\end{table}


\begin{figure*}[htb]
	\centering
	\subfigure[Random Forest (ADTM)]{
		\scalebox{0.23}[0.23]{
			\includegraphics[width=1\linewidth]{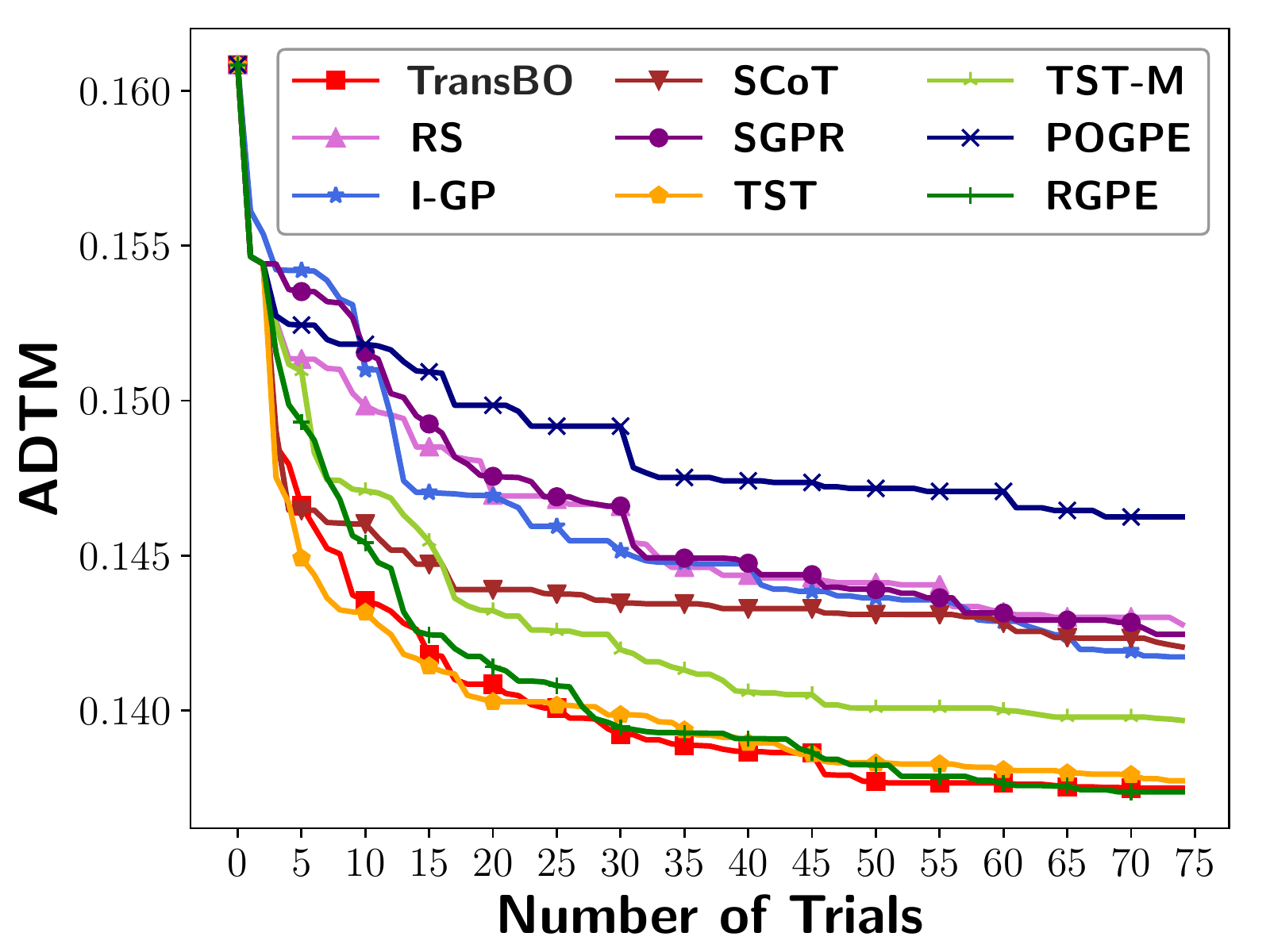}
	}}
	\subfigure[LightGBM (ADTM)]{
		\scalebox{0.23}[0.23]{
			\includegraphics[width=1\linewidth]{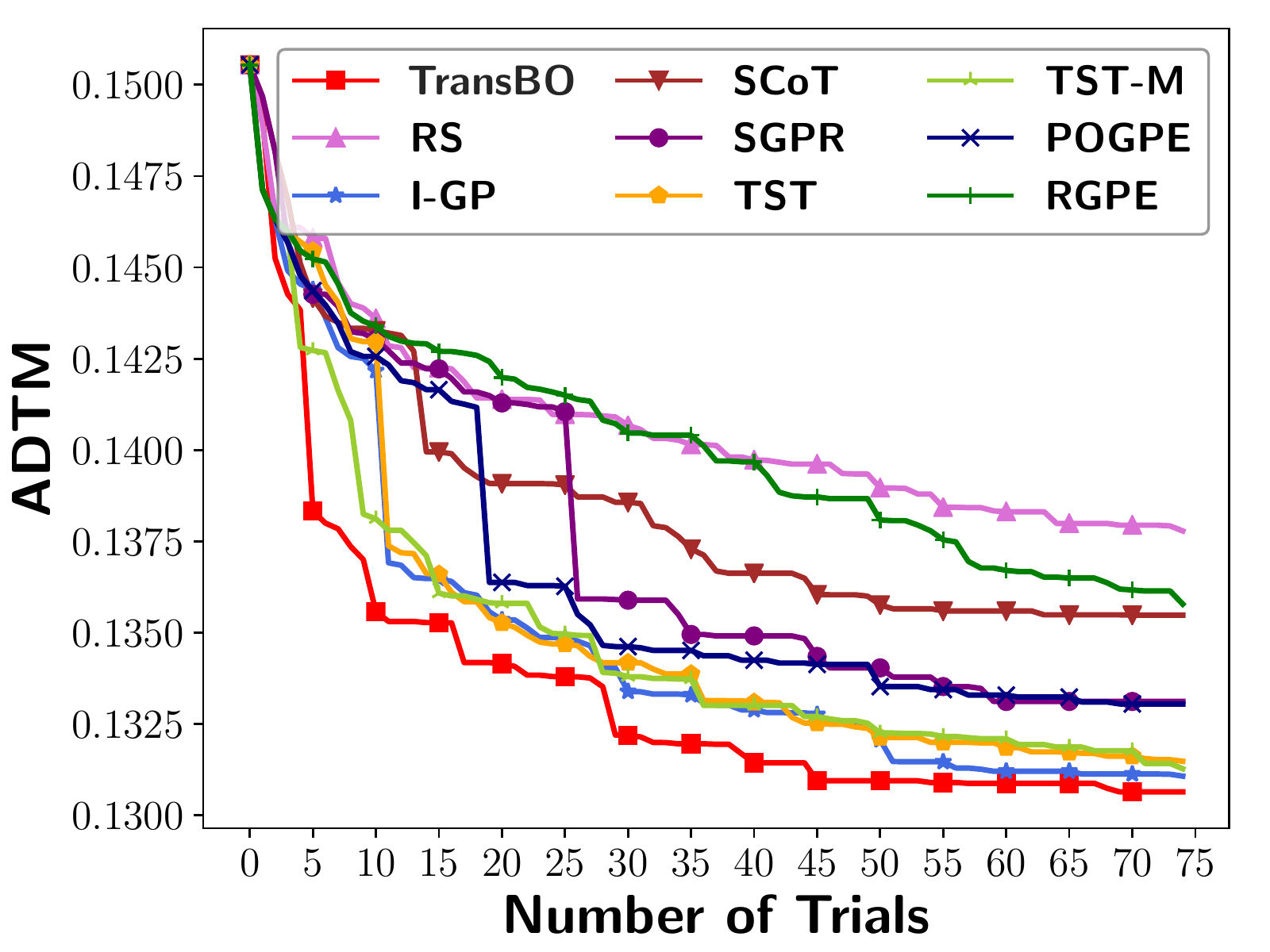}
	}}
	\subfigure[Adaboost (ADTM)]{
		\scalebox{0.23}[0.23]{
			\includegraphics[width=1\linewidth]{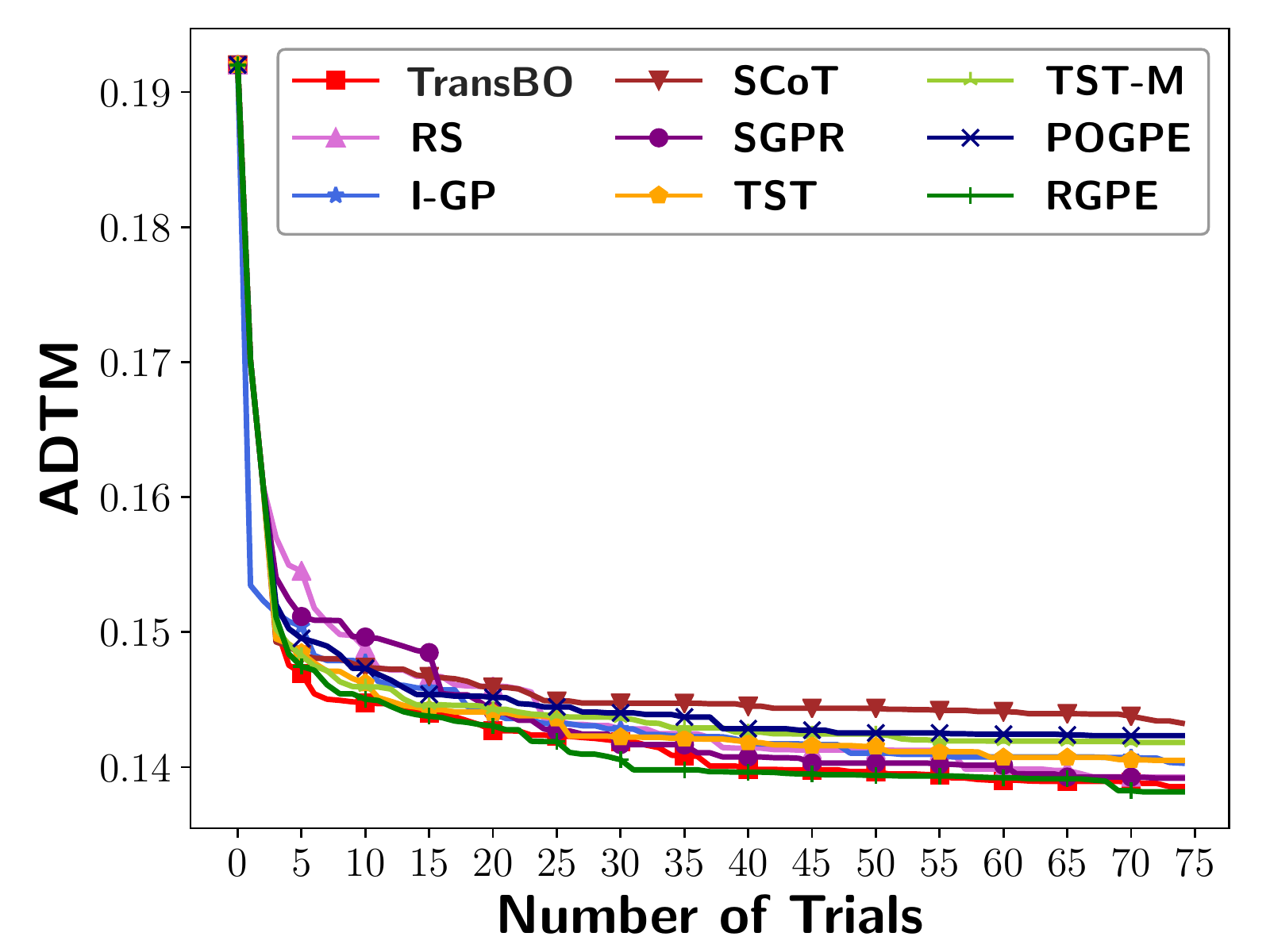}
	}}
	\subfigure[Extra Trees (ADTM)]{
		\scalebox{0.23}[0.23]{
			\includegraphics[width=1\linewidth]{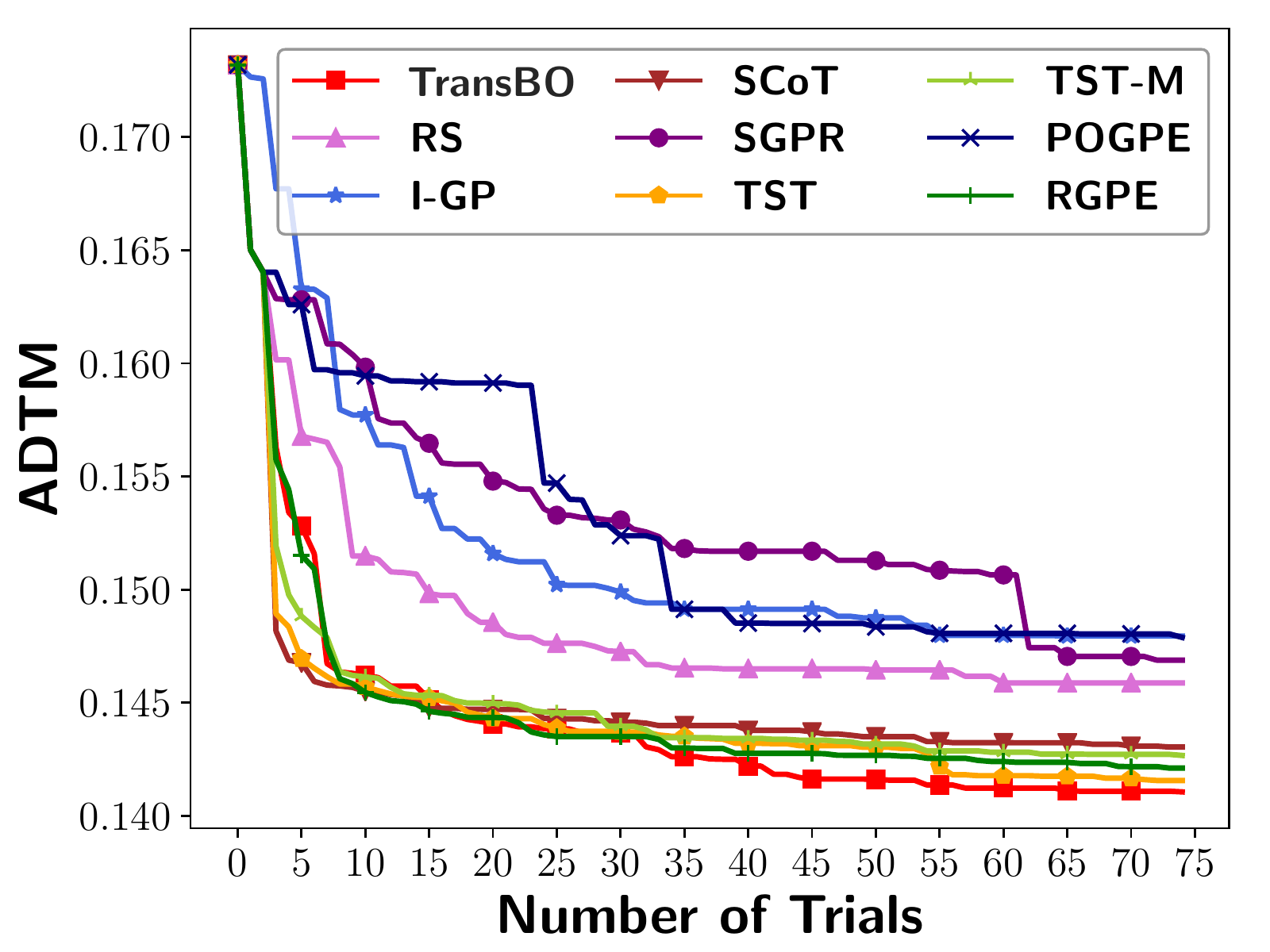}
	}}
	\vskip -0.1in
	\caption{Static results on four benchmarks with 75 trials.}
	\vskip -0.15in
  \label{offline_exp1_adtm}
\end{figure*}

\begin{figure*}[htb]
	\centering
	\subfigure[Random Forest (Average Rank)]{
		\scalebox{0.23}[0.23]{
			\includegraphics[width=1\linewidth]{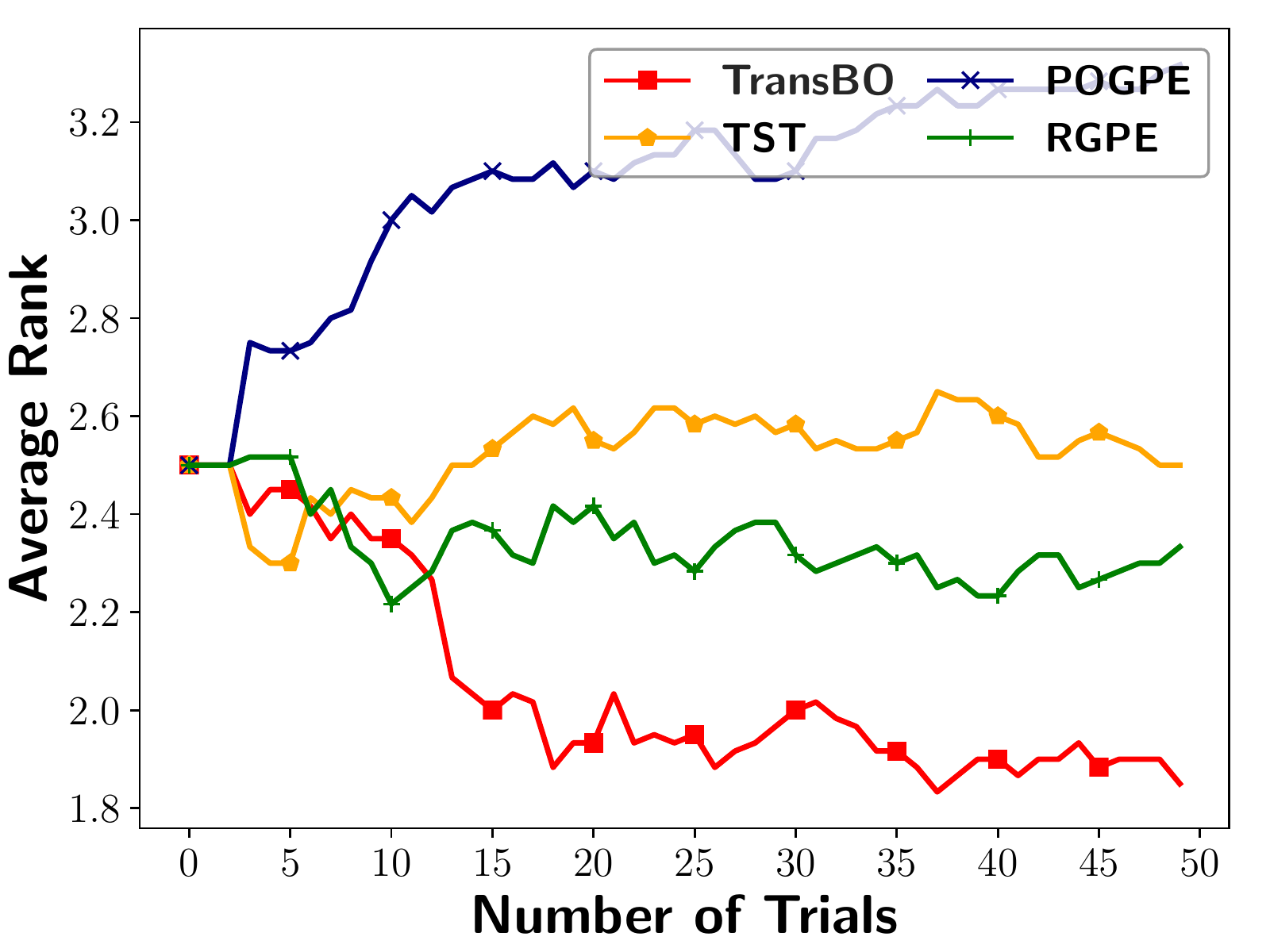}
	}}
	\subfigure[Random Forest (ADTM)]{
		\scalebox{0.23}[0.23]{
			\includegraphics[width=1\linewidth]{images/source_extraction/exp3_adaboost_50_adtm_result.pdf}
	}}
	\subfigure[Extra Trees (Average Rank)]{
		\scalebox{0.23}[0.23]{
			\includegraphics[width=1\linewidth]{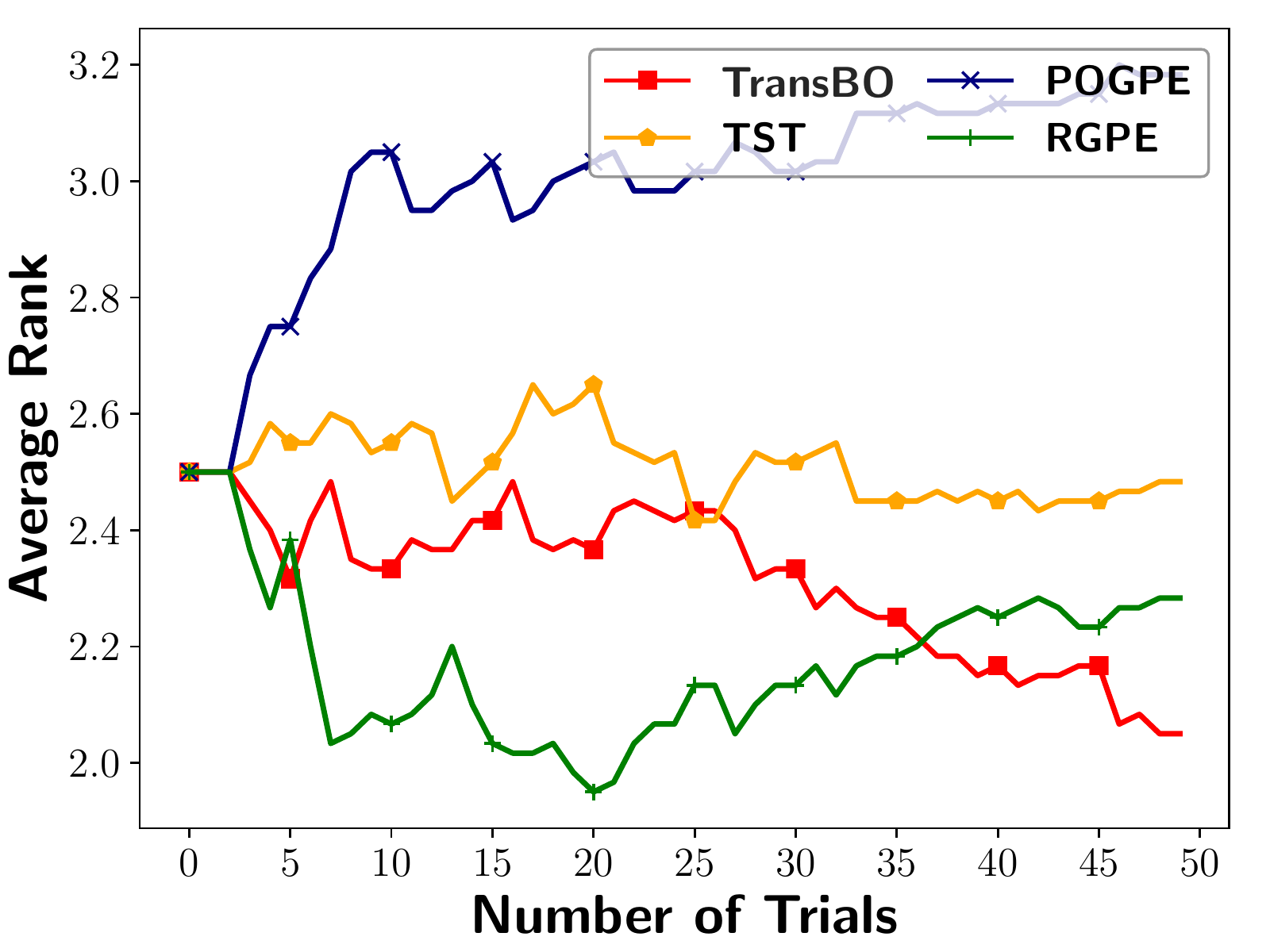}
	}}
    \subfigure[Extra Trees (ADTM)]{
		\scalebox{0.23}[0.23]{
			\includegraphics[width=1\linewidth]{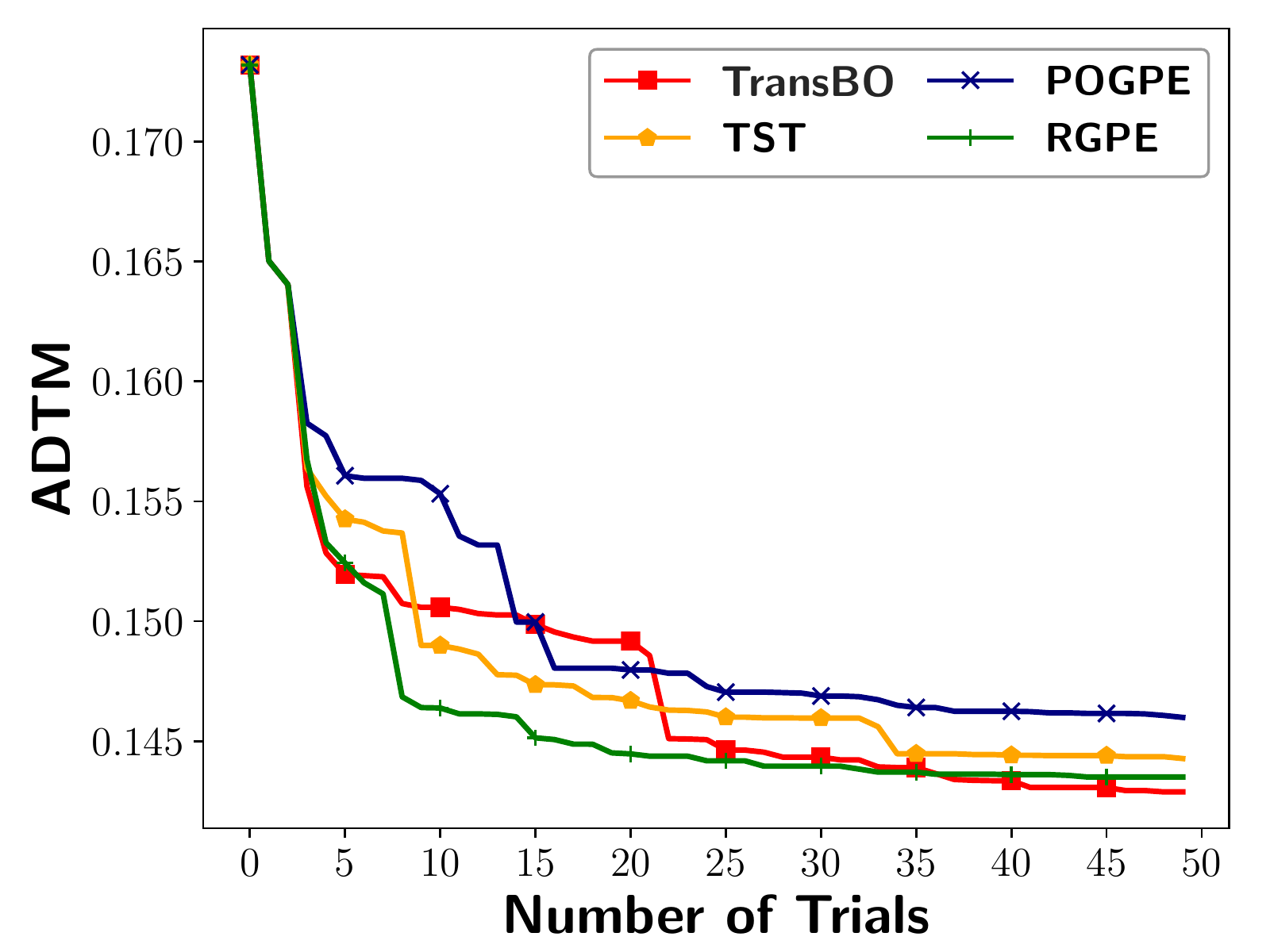}
	}}
	\vskip -0.1in
	\caption{Results on source knowledge learning.}
  \label{source_ext_exp3_2}
\end{figure*}

\begin{table}[h]
\centering
\small
\resizebox{0.8\columnwidth}{!}{
\begin{tabular}{cccc}
    \toprule
    Name & \#Rows & \#Columns & \#Categories \\ 
    \midrule
    balloon & 2001 & 1 & 2 \\
    kc1 & 2109 & 21 & 2 \\
    quake & 2178 & 3 & 2 \\
    segment & 2310 & 19 & 7 \\
    madelon & 2600 & 500 & 2 \\
    space\_ga & 3107 & 6 & 2 \\
    splice & 3190 & 60 & 3 \\
    kr-vs-kp & 3196 & 36 & 2 \\
    sick & 3772 & 29 & 2 \\
    hypothyroid(1) & 3772 & 29 & 4 \\
    hypothyroid(2) & 3772 & 29 & 2 \\
    pollen & 3848 & 5 & 2 \\
    analcatdata\_supreme & 4052 & 7 & 2 \\
    abalone & 4177 & 8 & 26 \\
    spambase & 4600 & 57 & 2 \\
    winequality\_white & 4898 & 11 & 7 \\
    waveform-5000(1) & 5000 & 40 & 3 \\
    waveform-5000(2) & 5000 & 40 & 2 \\
    page-blocks(1) & 5473 & 10 & 5 \\
    page-blocks(2) & 5473 & 10 & 2 \\
    optdigits & 5610 & 64 & 10 \\
    satimage & 6430 & 36 & 6 \\
    wind & 6574 & 14 & 2 \\
    musk & 6598 & 167 & 2 \\
    delta\_ailerons & 7129 & 5 & 2 \\
    mushroom & 8124 & 22 & 2 \\
    puma8NH & 8192 & 8 & 2 \\
    cpu\_small & 8192 & 12 & 2 \\
    cpu\_act & 8192 & 21 & 2 \\
    bank32nh & 8192 & 32 & 2 \\
    \bottomrule
\end{tabular}
}
\caption {Details of 30 datasets used in the benchmarks.}
\label{cls_datasets}
\end{table}

\begin{figure} 
    \subfigure[quake]{
     \begin{minipage}[h]{0.46\linewidth}
        \centering   
        \includegraphics[width=1.\linewidth]{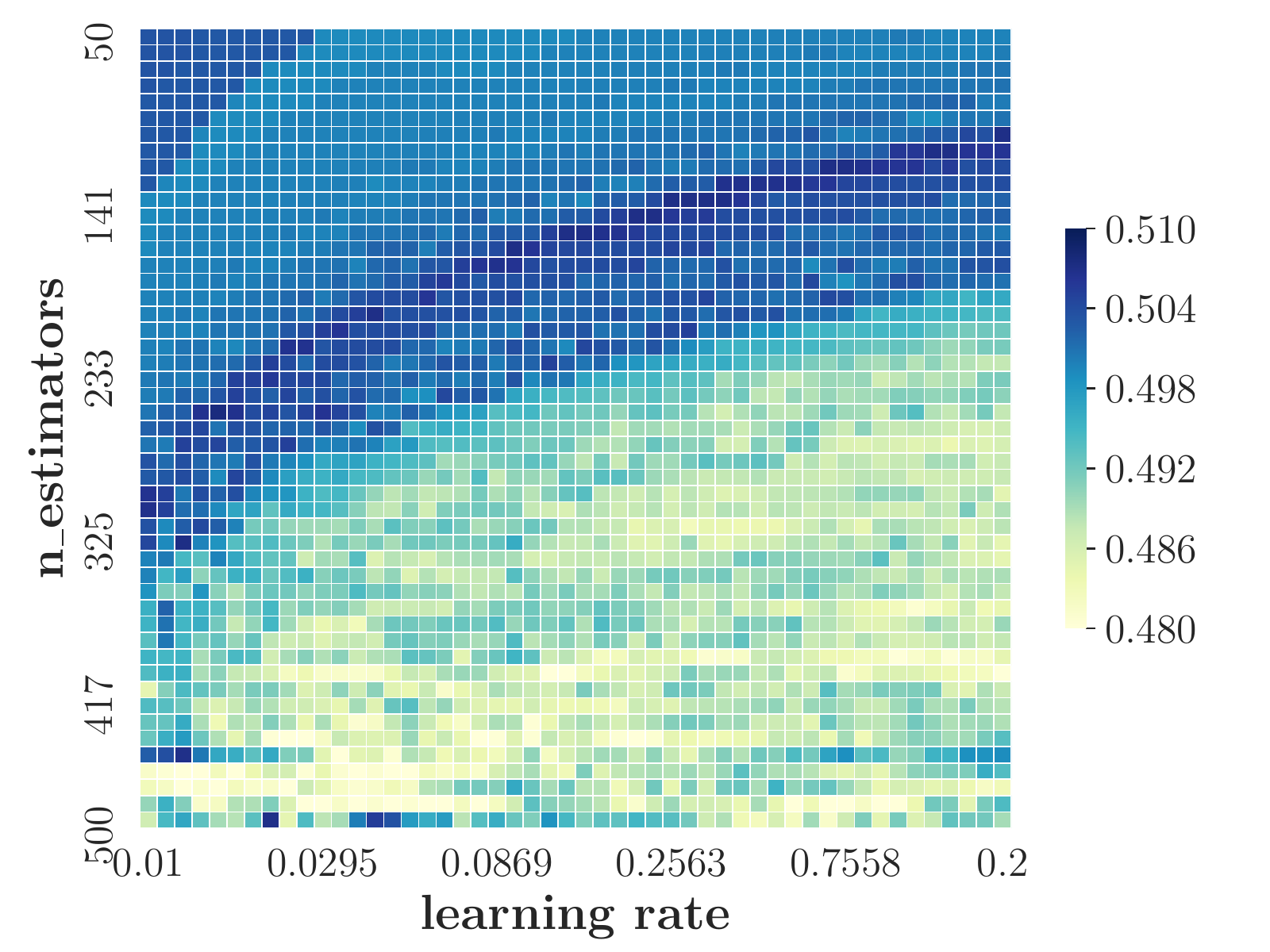}
        \vskip -0.15in
     \end{minipage}
    }
  \subfigure[hypothyroid(2)]{
     \begin{minipage}[h]{0.46\linewidth}   
        \centering
        \includegraphics[width=1.\linewidth]{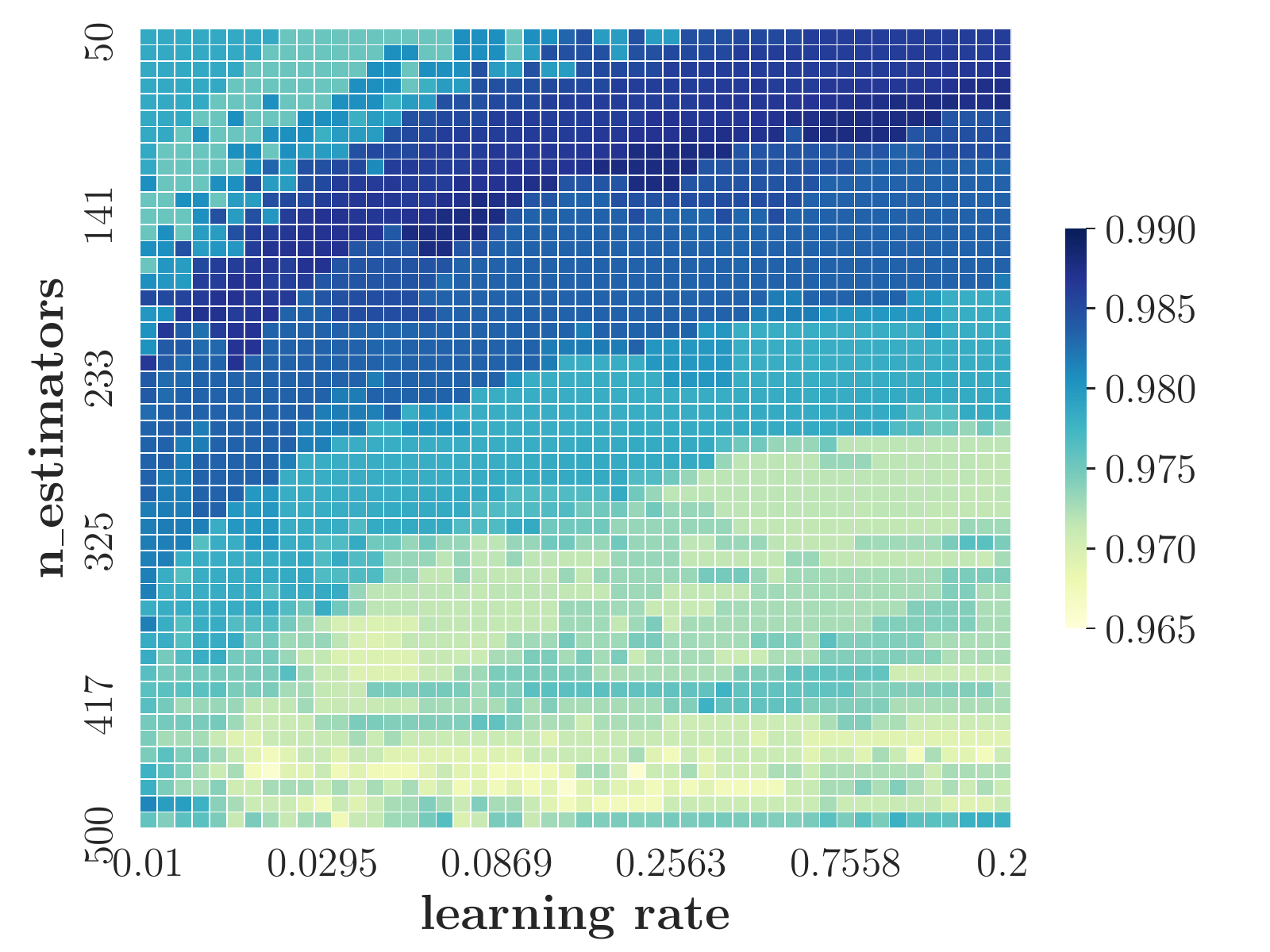}
        \vskip -0.15in
     \end{minipage}
    }
    \vskip -0.15in
    \caption{Performance of 2500 Adaboost configurations on two tasks, in which each hyperparameter has 50 settings.}
    \label{task_heatmap}
    \vskip -0.1in
\end{figure}

\subsection{Feasibility of Transfer Learning}
\label{a.2}
To verify the feasibility of transfer learning in the setting of HPO, we conduct an HPO experiment on two datasets --- quake and hypothyroid(2). 
We tune the learning rate and n\_estimators of Adaboost while fixing the other hyperparameters, and then evaluate the validation performance (the balanced accuracy) of each configuration. 
Figure~\ref{task_heatmap} shows the performance on 2500 Adaboost configurations, where deeper color means better performance. 

It is quite clear that the optimal configuration differs on the two datasets (tasks), which means re-optimization is essential for HPO. 
However, the performance distribution is somehow similar on the two datasets. 
For example, they both perform badly in the lower-right region and perform well in the upper region. 
Based on this observation, it is natural to accelerate the re-optimization process with the auxiliary knowledge acquired from the previous tasks.

\subsection{More Experiment Results}
\label{a.3}
In this section, we provide more experiment results besides those in Section~\ref{sec:exp_sec}.

\para{Static Experiments}
Figure~\ref{offline_exp1_adtm} shows the results of all considered methods on the four benchmarks, where the metric is ADTM.
We can observe that the proposed \sys exhibits strong stability, and performs well across benchmarks.

\para{Source Knowledge Learning}
The additional results on Random Forest and Extra Trees are illustrated in Figure~\ref{source_ext_exp3_2}. 
Similar to the findings in Section~\ref{sec:abla}, 
our method - \sys shows excellent ability in extracting and integrating the source knowledge from previous tasks.

\subsection{Convergence Discussion about \sys}
\label{converge_analysis}
In \sys, when sufficient trials on the target task are obtained, the weight of target surrogate $p^{T}$ will approach 1 as the HPO proceeds. 
Based on the mechanism we adopted in \sys --- cross-validation, we can observe that $p^{T}_{i}$ in the $i$-th trial will approach 1 as $i$ increases. 
Therefore, the final TL surrogate $M^{TL}$ will be set to the target surrogate $M^{T}$.
So we can have that, \newline
\emph{With sufficient trials, the final TL surrogate will find the same optimum as the target surrogate does; that's, the final solution of surrogate $M^{TL}$ will be no worse than the one in $M^{T}$ given sufficient trials.} \newline
The above finding demonstrates that \sys can alleviate negative transfer~\cite{pan2010survey}. In other words, it can avoid performance degradation compared with non-transfer methods -- the traditional BO methods.

\section{Reproduction Details}
\label{reproduction}
The source code and the benchmark data are available in the compressed file {\em ``benchmark\_data\_and\_source\_code.zip''} on CMT3. 
The source code is also available in the anonymous repository~\footnote{https://anonymous.4open.science/r/TransBO-EE01/} now.
All files in the benchmark should be placed under the folder {\em `data/hpo\_data'} of the project root directory.
To reproduce the experimental results in this paper, an environment of Python 3.6+ is required. We introduce the experiment scripts and installation of required tools in \emph{README.md} and list the required Python packages in \emph{requirements.txt} under the root directory. 
Take one experiment as an example, to evaluate the static TL performance of \sys and other baselines on Random Forest using 29 source tasks with 75 trials, you need to execute the following script: \newline
{\em
python tools/static\_benchmark.py --trial\_num 75 --algo\_id random\_forest --methods rgpe,pogpe,tst,transbo --num\_source\_problem 29
}

Please check the document \emph{README.md} in this repository for more details, e.g., how to use the benchmark, and how to run the other experiments.

\end{document}